\documentclass{article}
\usepackage{times}
\usepackage{helvet}
\usepackage{courier}
\usepackage{float}
\usepackage[letterpaper, margin=1in]{geometry}
\usepackage{listings}
\usepackage{nicefrac}

\usepackage{hyperref}
\hypersetup{plainpages=false,
	breaklinks=true,
	colorlinks=true,
	urlcolor=blue,
	citecolor=blue,                                       
	linkcolor=blue,
	bookmarksopen=true,
	bookmarksdepth=3,
	bookmarksopenlevel=3,
	pdfstartview=FitV,
	pdfdisplaydoctitle=true}

\usepackage{natbib}
\usepackage[american]{babel}

\usepackage[utf8]{inputenc}
\usepackage{csquotes}
\usepackage{amsmath}
\usepackage{amsthm}
\usepackage{amsfonts}
\usepackage{graphicx}
\usepackage[colorinlistoftodos]{todonotes}
\usepackage{afterpage}
\usepackage{comment}
\usepackage{subcaption}
\usepackage{algorithm}
\usepackage{algorithmic}
\usepackage[title]{appendix}

\DeclareGraphicsExtensions{.pdf, .png, .jpg}

\newcommand{\states}{\mathcal{S}}
\newcommand{\actions}{\mathcal{A}}
\newcommand{\reals}{\rm I\!R}
\newcommand{\tr}{^\top}
\newcommand{\expect}{\mathbb{E}}

\newcommand{\x}{\mathbf{\phi}}
\newcommand{\e}{\mathbf{\nu}}
\newcommand{\w}{\mathbf{w}}
\newcommand{\Cumulant}{C}

\newcommand{\target}{k}
\newcommand{\Kset}{\Gamma}
\newcommand{\kset}{\gamma_\target}

\newcommand{\plusequals}{\mathrel{+}=}

\newcommand{\imgheight}{2.2in}

\newcommand{\thename}{$\Gamma$-net}
\newcommand{\thenames}{{\thename s}}

\title{Gamma-Nets: Generalizing Value Estimation Over Timescale}
\author{ \textbf{Craig Sherstan,\textsuperscript{\rm 1} 
		Shibhansh Dohare,\textsuperscript{\rm 1}}\\  
		\textbf{James MacGlashan,\textsuperscript{\rm 3} 
		Johannes G\"unther,\textsuperscript{\rm 1} 
		Patrick M. Pilarski\textsuperscript{\rm 1,2}} \\ 
	\textsuperscript{\rm 1}Department of Computing Science, University of Alberta, Canada\\
	\textsuperscript{\rm 2}Department of Medicine, University of Alberta, Canada\\
	\textsuperscript{\rm 3}Cogitai, USA\\
	sherstan@ualberta.ca, pilarski@ualberta.ca 
}
\begin{document}
	
	\maketitle
\begin{abstract}
	Temporal abstraction is a key requirement for agents making decisions over long time horizons---a fundamental challenge in reinforcement learning. There are many reasons why making value estimates at multiple timescales might be useful; recent work has shown that value estimates at different time scales can be the basis for creating more advanced discounting functions and for driving representation learning. Further, predictions at many different timescales serve to broaden an agent's model of its environment. One predictive approach of interest within an online learning setting is general value function (GVFs), which represent models of an agent's world as a collection of predictive questions each defined by a policy, a signal to be predicted, and a prediction timescale. In this paper we present {\thenames}, a method for generalizing value function estimation over timescale, allowing a given GVF to be trained and queried for arbitrary timescales so as to greatly increase the predictive ability and scalability of a GVF-based model. The key to our approach is to use timescale as one of the value estimator’s inputs. As a result, the prediction target for any timescale is available at every timestep and we are free to train on any number of timescales. We first provide two demonstrations by 1) predicting a square wave and 2) predicting sensorimotor signals on a robot arm using a linear function approximator. Next, we empirically evaluate {\thenames} in the deep reinforcement learning setting using policy evaluation on a set of Atari video games. Our results show that {\thenames} can be effective for predicting arbitrary timescales, with only a small cost in accuracy as compared to learning estimators for fixed timescales. {\thenames} provide a method for accurately and compactly making predictions at many timescales without requiring a priori knowledge of the task, making it a valuable contribution to ongoing work on model-based planning, representation learning, and lifelong learning algorithms.
\end{abstract}	
	
\section{Value Functions and Timescale}
Reinforcement learning (RL) studies algorithms in which an agent learns to maximize the amount of reward it receives over its lifetime. A key method in RL is the estimation of \textit{value}---the expected cumulative sum of discounted future rewards (called the \textit{return}). In loose terms this tells an agent how good it is to be in a particular state. The agent can then use value estimates to learn a \textit{policy}---a way of behaving---which maximizes the amount of reward received.

\citet{Sutton2011} broadened the use of value estimation by introducing general value functions (GVFs), in which value estimates are made of other sensorimotor signals, not just reward. GVFs can be thought of as representing an agent's model of itself and its environment as a collection of questions about future sensorimotor returns; a predictive representation of state \citep{Dayan1993}.
A GVF is defined by three elements: 1) the policy, 2) the \textit{cumulant} (the sensorimotor signal to be predicted), and 3) the prediction timescale, $\gamma$. Considering a simple mobile robot, examples of GVF questions include ``How much current will my motors consume over the next 3 seconds if I spin clockwise?'' or ``How long until my bump sensor goes high if I drive forward?''

\begin{figure}[t]
	\begin{center}
		\centerline{\includegraphics[width=0.95\columnwidth]{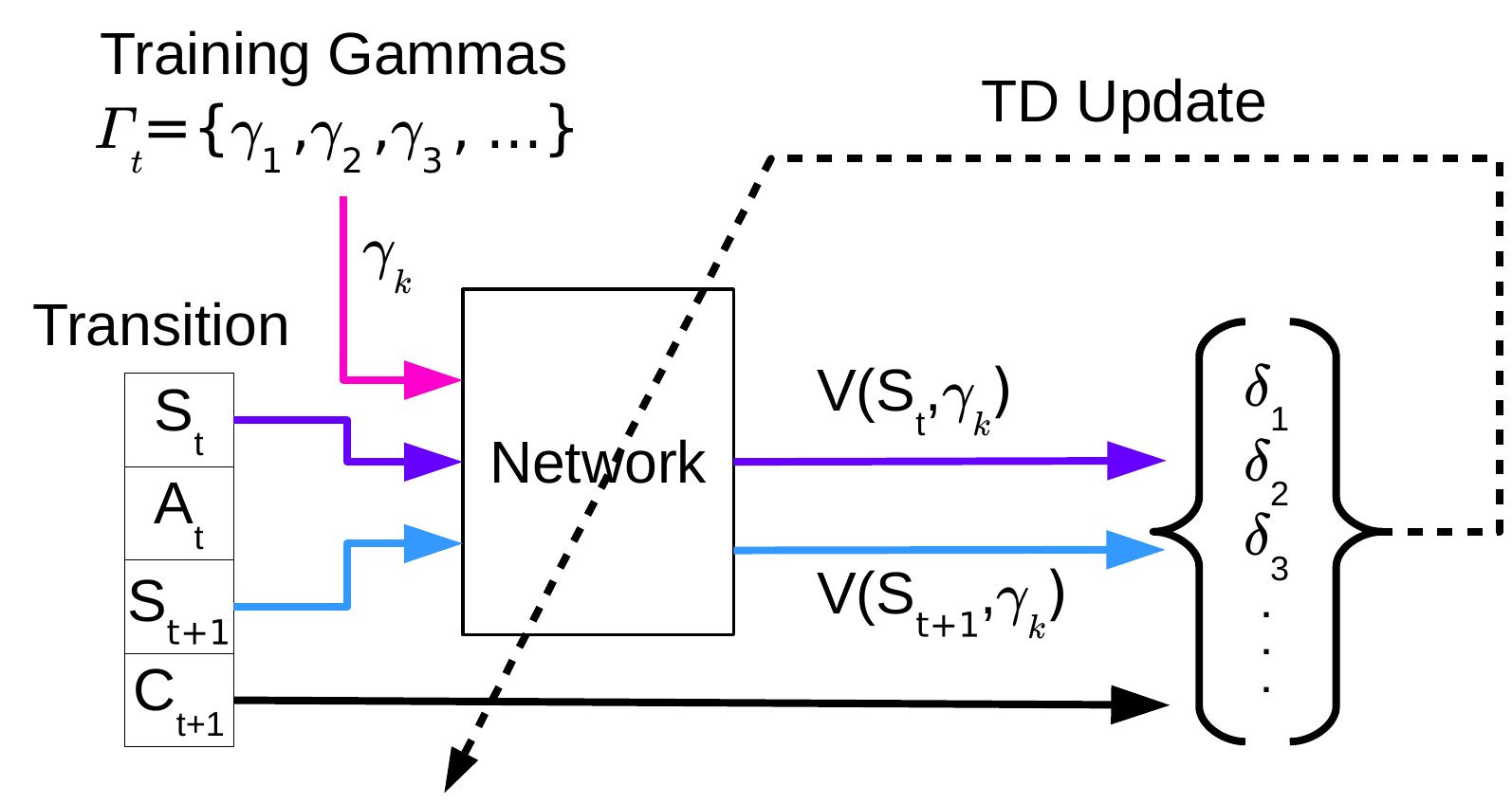}}
		\caption{\textbf{Training $\boldsymbol{\gamma}$-nets}. Values are estimated by providing state and timescale, $\gamma$, as inputs to the network parameterized by weights $\w$. An agent in state $S$ takes action $A$ and transitions to state $S'$ receiving the new target signal $\Cumulant$. The agent selects a set of timescales $\Gamma_t$ on which to train and for each $\gamma_k\in\Gamma_t$ computes values $V(S,\gamma_k;\w)$ and $V(S',\gamma_k;\w)$. For each $\gamma_k$, the TD error is calculated according to $\delta_k=\Cumulant + \gamma_k V(S',\gamma_k;\w) - V(S,\gamma_k;\w)$. The TD errors are then collected and used to update $\w$ using a chosen TD learning algorithm, such as TD($\lambda$) or GTD.}
		\label{fig:alg}
	\end{center}
\end{figure}

Modeling the world at many timescales is seen as a key problem in artificial intelligence \citep{Sutton1995,Sutton1999a}. Further, there is evidence that humans and other animals make estimates of reward and other signals at numerous timescales \citep{Tanaka2016}. This paper focuses on generalizing value estimation over timescale. Our work can be seen as directly connected to the concept of \textit{nexting}, in which animals and people make large numbers of predictions of sensory input at many, short-term, timescales \citep{Gilbert2006}. \citet{Modayil2014a} demonstrated the concept of nexting using GVFs on a mobile robot. Until now, value estimation has generally been limited to a single fixed timescale. That is, for each desired timescale, a discrete and unique predictor was learned. However, there are situations where we may desire to have value estimates of the same cumulant over many different timescales. For example, consider an agent driving a car. Such an agent may make numerous predictions about the likelihood of colliding with various objects in its vicinity. The agent needs to consider the risk of collisions in both the near term and far term and the relevance of each may change with the speed of the car. If the engineer knew which timescales would be needed ahead of time they could design them into the system, but this is not the case for complex settings.

Here we present a novel class of algorithms which enables the explicit learning and inference of value estimates for any valid fixed discount. The key insights to our approach are: 1) the timescale can be treated as an input parameter for inference and learning and 2) the estimated bootstrapped prediction target for any fixed timescale is available at every timestep. We demonstrate {\thenames} in three policy evaluation settings: 1) predicting a square wave, 2) predicting sensorimotor signals on a robot arm, 3) predicting reward in Atari video games.

The ideas behind our approach are based on work by \citet{Schaul2015} which generalized value estimation across goals by providing a goal embedding vector as input to the value network. In contrast, our approach provides the discount, $\gamma$ as input. \citet{Xu2018} also provide $\gamma$ as input to their value and policy networks. They present a meta-learning approach which learns the best $\gamma$ to provide to an inner policy. Here we focus on determining what is necessary to effectively train a value network to train over timescale. Additionally, our algorithm trains on multiple timescales simultaneously.
	
	
	\section{Background}
	\label{sec:background}
	We model the environment as a Markov Decision Process. At each timestep $t$ the agent, in state $S_t\in\states$, takes action $A_t\in\actions$ according to policy $\pi:\states\times\actions\rightarrow[0,1]$ and transitions to state $S_{t+1}\in\states$ according to the transition probability $p(\cdot | S_t,A_t)$. In the traditional RL setting the agent receives a reward $R_{t+1}\equiv R(S_t,A_t,S_{t+1})\in\reals$. The agent tries to learn a policy which maximizes the cumulative reward it receives in the future, which is defined as the return: $G_t=R_{t+1} + \gamma R_{t+2} + \gamma^2 R_{t+3}+\ldots$. In the case of GVFs we simply substitute our signal of interest, the cumulant, $\Cumulant$ for reward, $R$. The term $\gamma\in[0,1)$ is referred to by several names including the timescale, the continuation function and the discount; it represents the amount of emphasis applied to future rewards and is the focus of this paper.
	
	A value estimate is simply the expectation of the return: $V_\pi(s)=\expect_\pi\big[G_t|S_t=s\big]$.
	Temporal difference (TD) learning is a common class of algorithms used in RL for learning an approximation of value \citep{Sutton1998}. Estimation weights are typically trained by semi-gradient descent using the TD error: $\delta_{t} = \Cumulant_{t+1} + \gamma V(S_{t+1}) - V(S_{t})$.
	
	While simple domains can be represented using tabular lookup, complex settings in which the state space is very large or infinite must use function approximation (FA) methods to estimate the value as $V(s;\w)$, where $\w$ is a set of weights parameterizing the network. Function approximation has the advantage that states are not treated independently, but rather, a learning step updates related states as well, allowing for generalization across state-space.
	
	\section{Generalizing over Timescale}
	\label{sec:gamma-nets}
	
	Our goal is to be able to predict the value function for any discount factor $\gamma$. While the GVF specification allows for $\gamma$ that are a function of the transition, here we focus solely on the case of fixed timescale. To achieve that goal, we propose $\Gamma$-nets: an architecture for value functions that operates not only on the state, but also the desired target discount factor $\gamma_\target$ (see Figure~\ref{fig:alg}). On each transition the network is trained on many $\gamma_\target\in\Gamma_t$ values. Thus, the $\Gamma$-net learns to generalize over arbitrary $\gamma_\target$ values. 
	
	Generating the error function for a {\thename} is also straightforward. For any single $\gamma_\target\in\Gamma_t$, the TD error is:
	\begin{align}
	\delta_{t;\gamma_{\target}} = \Cumulant_{t+1} + \gamma_\target V(S_{t+1}, \gamma_\target) - V(S_{t}, \gamma_\target).
	\end{align}
	The total gradient can then be summed over all $\gamma_\target\in\Gamma_t$ and applied to update the network.
	
\begin{figure}[ht!]
		\begin{center}
			\centerline{\includegraphics[height=\imgheight]{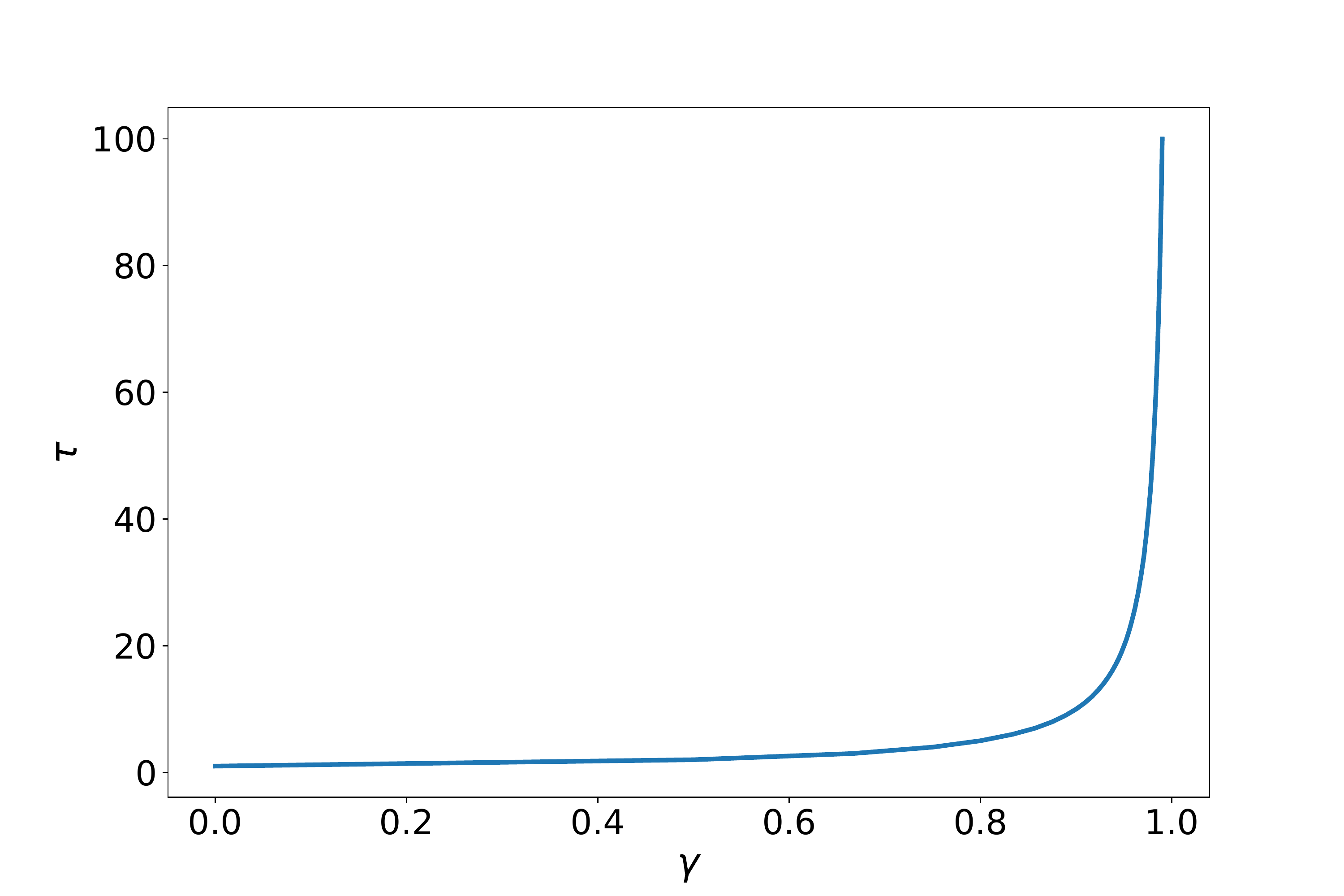}}
			\caption{Non-linear relationship between discount $\gamma$ and prediction length in expected timesteps (ts).}
			\label{fig:gamma_ts}
		\end{center}
	\end{figure}

	Choosing $\Gamma_t$ must be done with care. A naive approach might uniformly sample $\gamma_\target\in [0, 1)$. However, value functions change non-linearly with $\gamma$. To illustrate this property, consider that $\gamma$ can be viewed as the probability of continuation, allowing us to derive the expected number of timesteps (ts) until termination of the return as (see \citet{Sherstan2015b} for a derivation):
	\begin{align}
	\tau=\frac{1}{1-\gamma}.
	\label{eq:tau_gamma}
	\end{align}
	Table~\ref{table:expected_ts} shows selected values of $\gamma$ and their corresponding values of $\tau$. The relationship between $\gamma$ and $\tau$ is non-linear for large values of $\gamma$ (Figure~\ref{fig:gamma_ts}). Thus, naively drawing $\gamma_\target$ from a uniform distribution would tend to favor very short timescales. Conversely, drawing uniformly from $\tau$ would put little emphasis on short timescales. While the best method for selecting $\gamma_\target$ for training is outside the scope of this paper, we provide some comparisons in our experiments. Note that throughout this paper we will refer broadly to the word \textit{timescale} for which we will use the parameters $\gamma$ or $\tau$ as appropriate. It should be assumed that these terms can be used interchangeably using Eq.~\eqref{eq:tau_gamma}.

	\begin{table}[!ht]
		\caption{Expected Timesteps} \label{table:expected_ts}
		\begin{center}
			\begin{tabular}{|c|c|}
				\hline
				$\boldsymbol{\gamma}$ & $\boldsymbol{\tau}$\\
				\hline
				0 & 1 \\
				0.5 & 2 \\
				0.8 & 5 \\
				0.9 & 10 \\
				0.95 & 20 \\
				0.975 & 40 \\ 
				0.983 & 60 \\
				0.9875 & 80 \\
				0.99 & 100 \\
				\hline
			\end{tabular}
		\end{center}
	\end{table}
	
	The representation of timescale used for input to the network may affect the network's ability to represent different timescales. The $\gamma$ scale compresses long timescales but spreads short ones and in the $\tau$ scale we have the opposite effect. Thus, providing both $\gamma$ and $\tau$ as input may allow for good discrimination at all timescales.
	
	Finally, the magnitude of returns at different timescales can be very different. Larger returns can produce larger errors and corresponding larger gradients, which can effectively dominate the network weights. In general it is the longer timescales which will produce larger magnitude returns, but returns can be constructed for which the opposite is true. To prevent large magnitude returns from dominating the network weights we need to scale the returns in some way. We want to look for a general solution as we may not know beforehand which timescales are most important and thus seek a way to balance accuracy for all timescales. A general approach is given by \citet{VanHasselt2016}, in which they continually normalize the target to have a mean of 0 and variance of 1. This allows them to handle rewards of varying magnitude. Here, we take a simpler approach focusing on keeping the magnitude of the returns, as a function of timescale, in the same ballpark, by learning the value of a scaled cumulant: $f(s,\kset;\w)=\expect_{\pi}[\sum_{t=0}\kset^{t}(1-\kset)C_{t+1} | S_0=s]$. This will scale the loss by timescale and should result in smaller network weights. However, the resulting prediction must then be rescaled by dividing by $(1-\kset)$.  However, if we instead redefine our value estimator as
	
	\begin{align}
	V(s, \kset;\w)=\frac{f(s,\kset;\w)}{(1-\kset)}
	\label{eq:scaled_v}
	\end{align}
	
	then we can simply scale the TD loss by $(1-\kset)$. In the following we show this derivation for n-step returns. The TD error for the n-step scaled cumulant is:
	\begin{align*}
	\delta_{t;\gamma_{\target}} = (1-\gamma_\target)(\Cumulant_{t+1} + \gamma_{\target} \Cumulant_{t+2} + \ldots + \gamma_{\target}^{n-1}\Cumulant_{t+n})
	+ \gamma_{\target}^{n} f(S_{t+n}, \gamma_\target) - f(S_{t}, \gamma_\target).\\
	\end{align*}
	But, if we substitute in $f$ from Eq.~\ref{eq:scaled_v} we have:
	\begin{align*}
	=& (1-\gamma_\target)(\Cumulant_{t+1} + \gamma_{\target} \Cumulant_{t+2} + \ldots + \gamma_{\target}^{n-1}\Cumulant_{t+n}) 
	+ \gamma_{\target}^{n}(1-\gamma_\target)V(S_{t+n}, \gamma_\target) - (1-\gamma_\target)V(S_{t}, \gamma_\target)\\
	=& (1-\gamma_\target)(\Cumulant_{t+1} + \gamma_{\target} \Cumulant_{t+2} + \ldots + \gamma_{\target}^{n-1}\Cumulant_{t+n}
	+ \gamma_{\target}^{n}V(S_{t+n}, \gamma_\target) - V(S_{t}, \gamma_\target)).
	\end{align*}
	This results in scaled losses and gradients. This scaling can then be applied at either the loss or gradient level.
	
	\section{Experiments}
	\label{sec:exp}
	
	We first provide two proof of concept demonstrations using linear function approximation. The first on a square wave signal, which is easily understood. The second on a robot arm. Next we empirically evaluate {\thenames} in a deep learning setting by looking at performance on Atari games.
	
	\subsection{Square-wave}
	
	\begin{figure}[ht!]
		\begin{center}
			\centerline{\includegraphics[height=\imgheight]{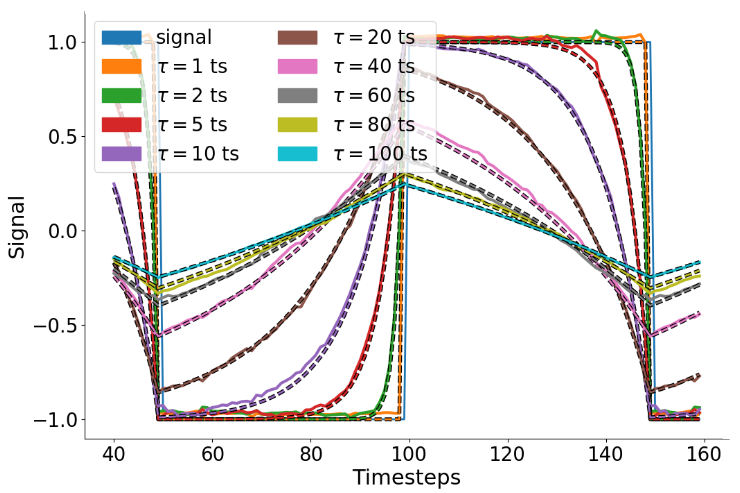}}
			\caption{\textbf{Square-wave Predictions.} Predictions (solid) against the true return (dashed) after 50k ts of training using both $\gamma$ and $\tau$ as input, scaling the loss and drawing two $\kset$ each from $\gamma$ and $\tau$ scales plus $\tau={1,100}$ . For display purposes all predictions are normalized by $(1-\gamma)$. We see good accuracy across all timescales.}
			\label{fig:simple_sq_scaled}
		\end{center}
	\end{figure}
	
	Our target signal was a repeating square wave 100 timesteps in length with a magnitude of $\{-1,1\}$ (Figure~\ref{fig:simple_sq_scaled}). Inputs were normalized and then tilecoded \citep{Sutton1998} with 20 tilings of width 1.0, 20 tilings of width 0.5 and 30 tilings of width 0.1. Tiling positions were randomly shifted by small amounts at the time of initialization for each run. Value estimates were computed using linear function approximation on the output of the tilecoding and the final layer of weights was updated using TD(0) \citep{Sutton1998} (The algorithm used for this experiment is given in Algorithm~\ref{alg:linear_TD0}). We also evaluated the impact of loss scaling. Unless otherwise stated: 1) timescale inputs were given on both the $\gamma$ and $\tau$ scales simultaneously, 2) $\Kset_t$ was 6 elements long, with $\tau\in\{1,100\}$ always included and two additional timescales drawn uniformly from each of the $\gamma$ and $\tau$ timescales, 3) loss scaling was used. Results are shown in Figure~\ref{fig:square_exp}. Each training run lasted for 50k timesteps and for each series 100 different runs were made. We show the normalized errors as a function of the prediction timescale, given on the $\tau$-scale. Results are averaged over the last 5k timesteps. For each $\tau$ we normalize by the maximum mean error across the series in the plot.

	\newcommand{\sqimgheight}{1.8in}
	\newcommand{\wmult}{0.45}
	\newlength{\tempdimb}
	\settoheight{\tempdimb}{\includegraphics[height=\sqimgheight]{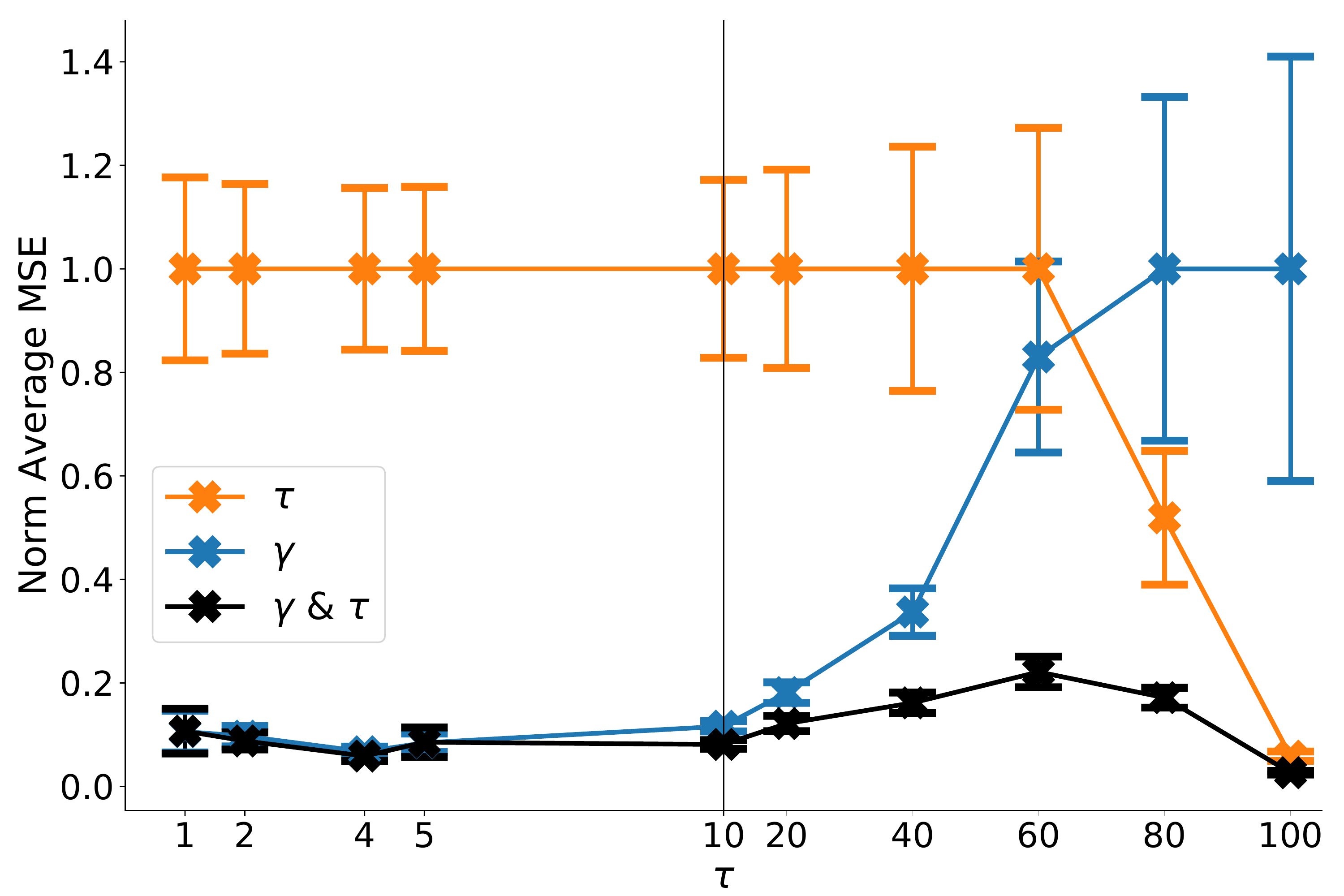}}
	\newcommand{\rownameb}[1]{
		\rotatebox{90}{\makebox[\sqimgheight][c]{\textbf{#1}}}
	}
	
	\begin{figure*}[t!]
		\centering
		\rownameb{Inputs}
		\begin{subfigure}[t]{\wmult\linewidth}
			\centering
			\includegraphics[height=\sqimgheight]{img/p_e/square/compare_inputs/MSE.pdf}
			\label{fig:sq:compare_inputs}
		\end{subfigure}
		\hfill
		\begin{subfigure}[t]{\wmult\linewidth}
			\centering
			\includegraphics[height=\sqimgheight]{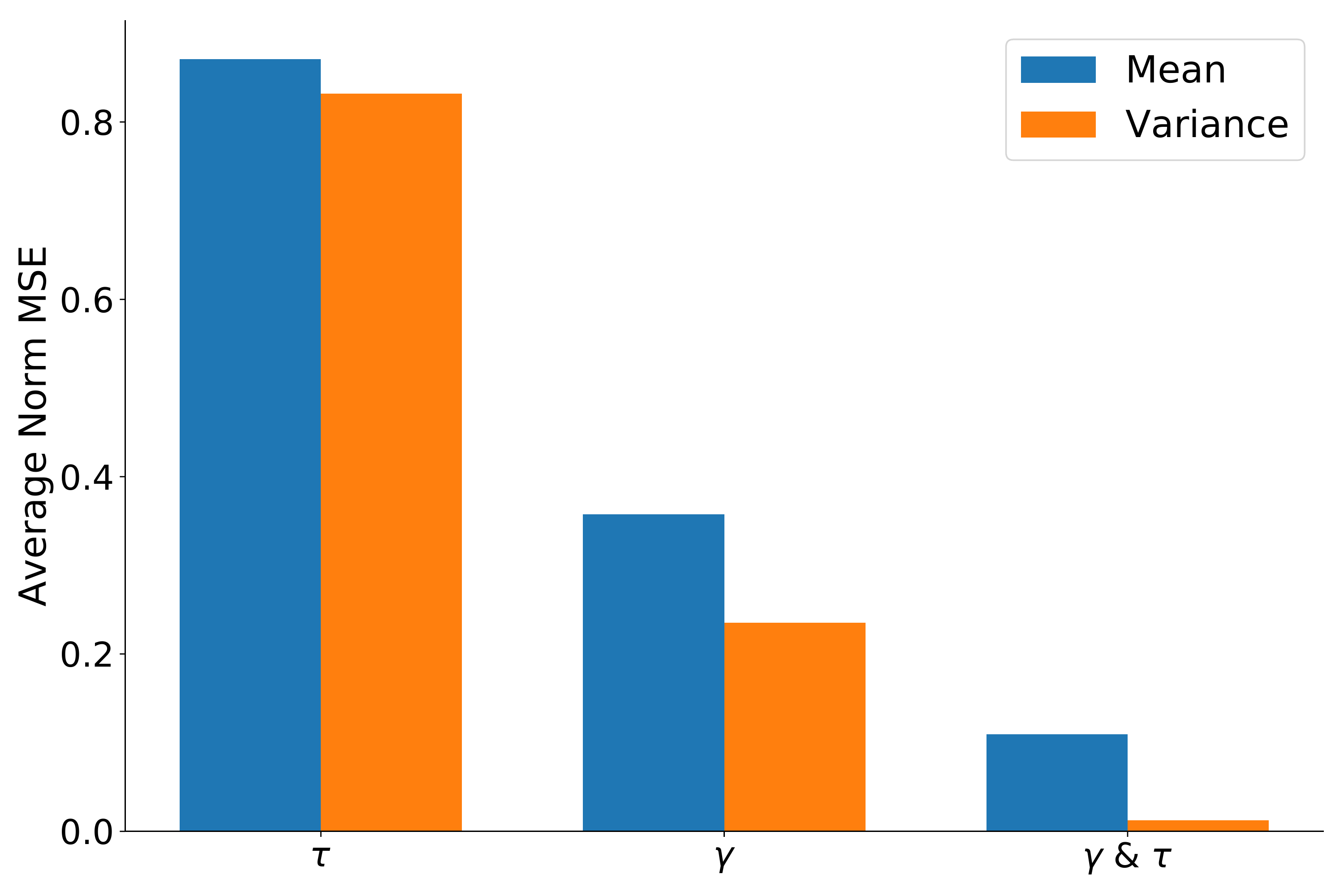}
			\label{fig:sq:compare_inputs_bar}
		\end{subfigure}
		\\
		\rownameb{Distribution}
		\begin{subfigure}[t]{\wmult\linewidth}
			\centering
			\includegraphics[height=\sqimgheight]{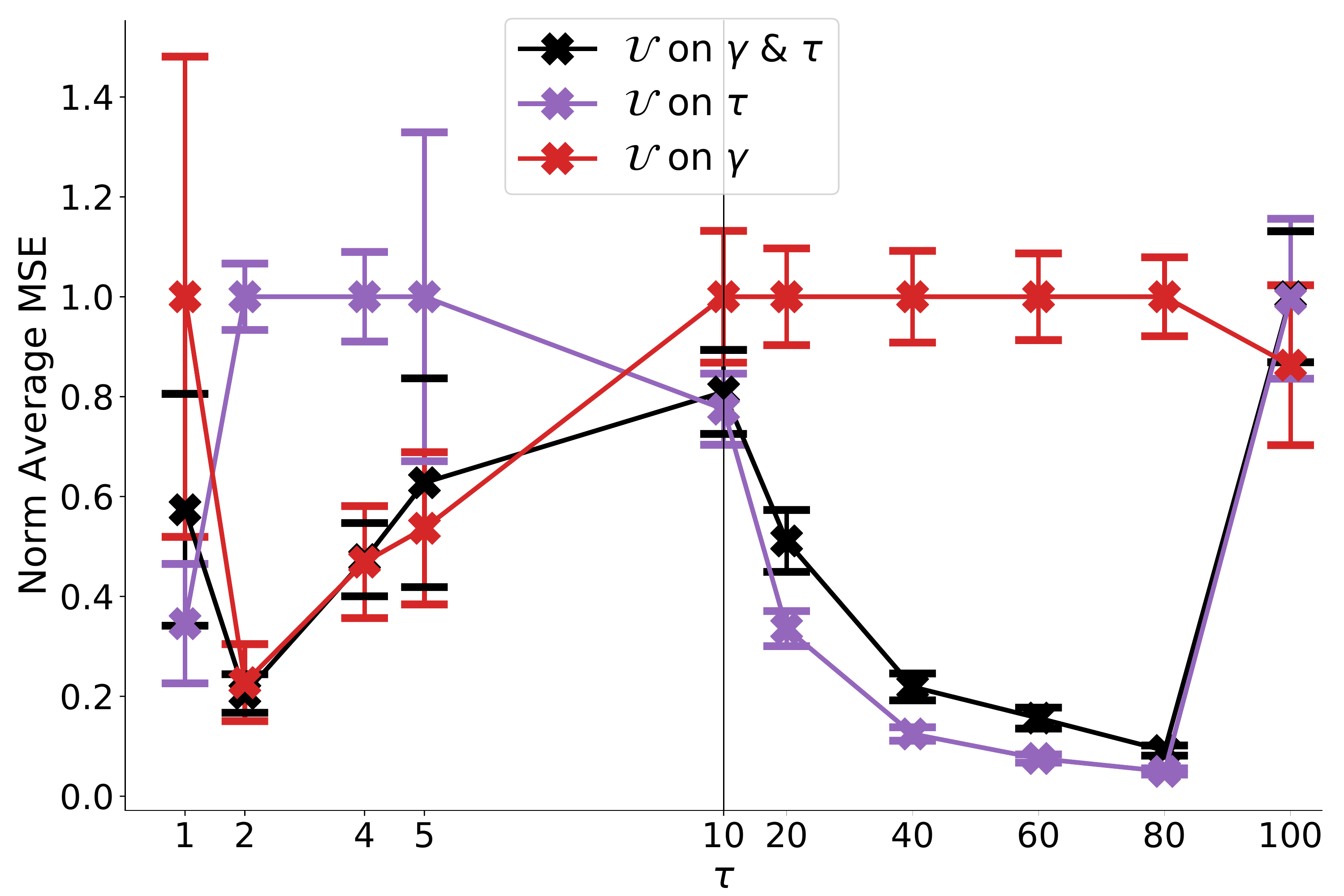}
			\label{fig:sq:compare_dist}
		\end{subfigure}
		\hfill
		\begin{subfigure}[t]{\wmult\linewidth}
			\centering
			\includegraphics[height=\sqimgheight]{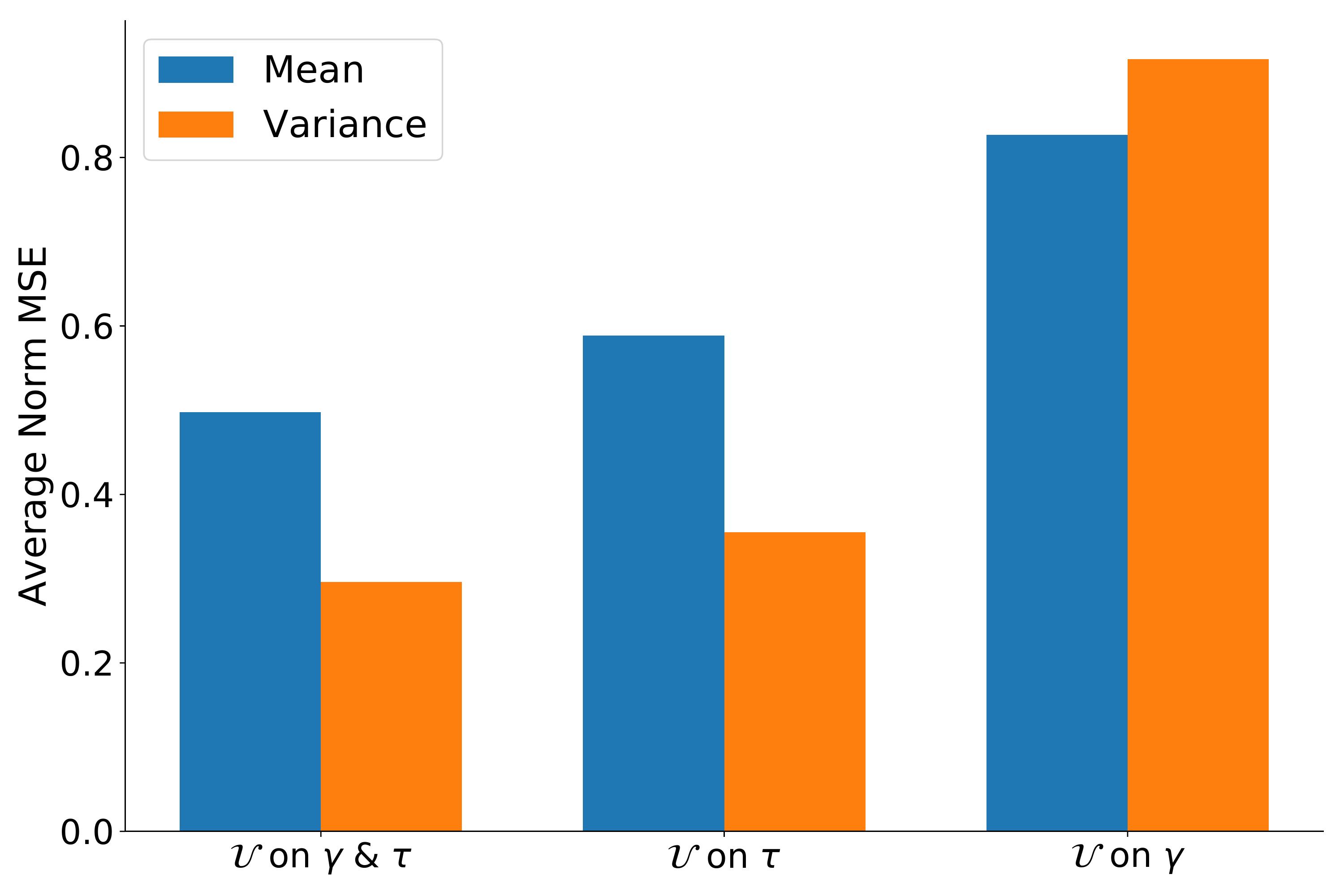}
			\label{fig:sq:compare_dist_bar}
		\end{subfigure}
		\\
		\rownameb{Scaling}
		\begin{subfigure}[t]{\wmult\linewidth}
			\centering
			\includegraphics[height=\sqimgheight]{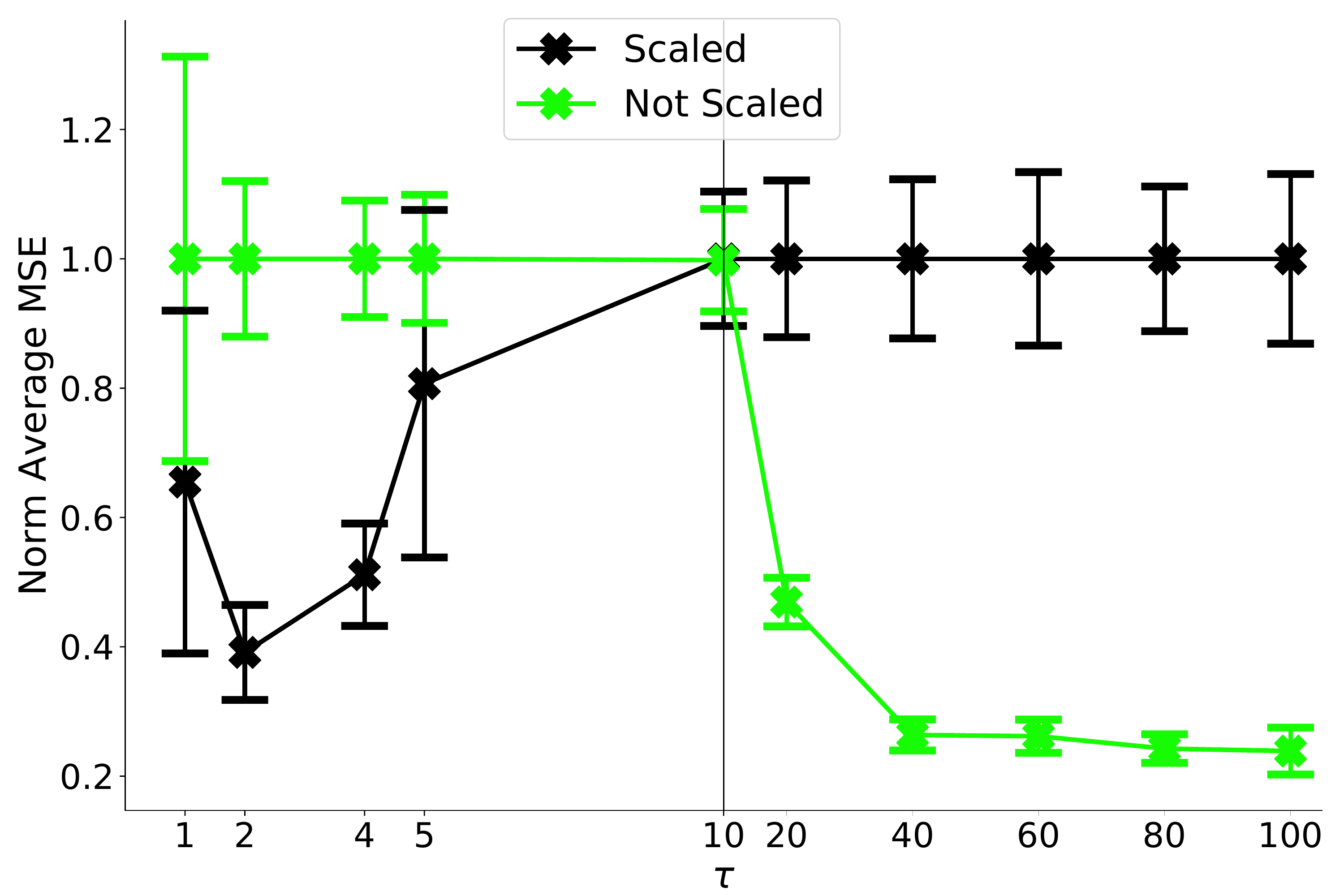}
			\caption{MSE}
			\label{fig:sq:compare_scale}
		\end{subfigure}
		\hfill
		\begin{subfigure}[t]{\wmult\linewidth}
			\centering
			\includegraphics[height=\sqimgheight]{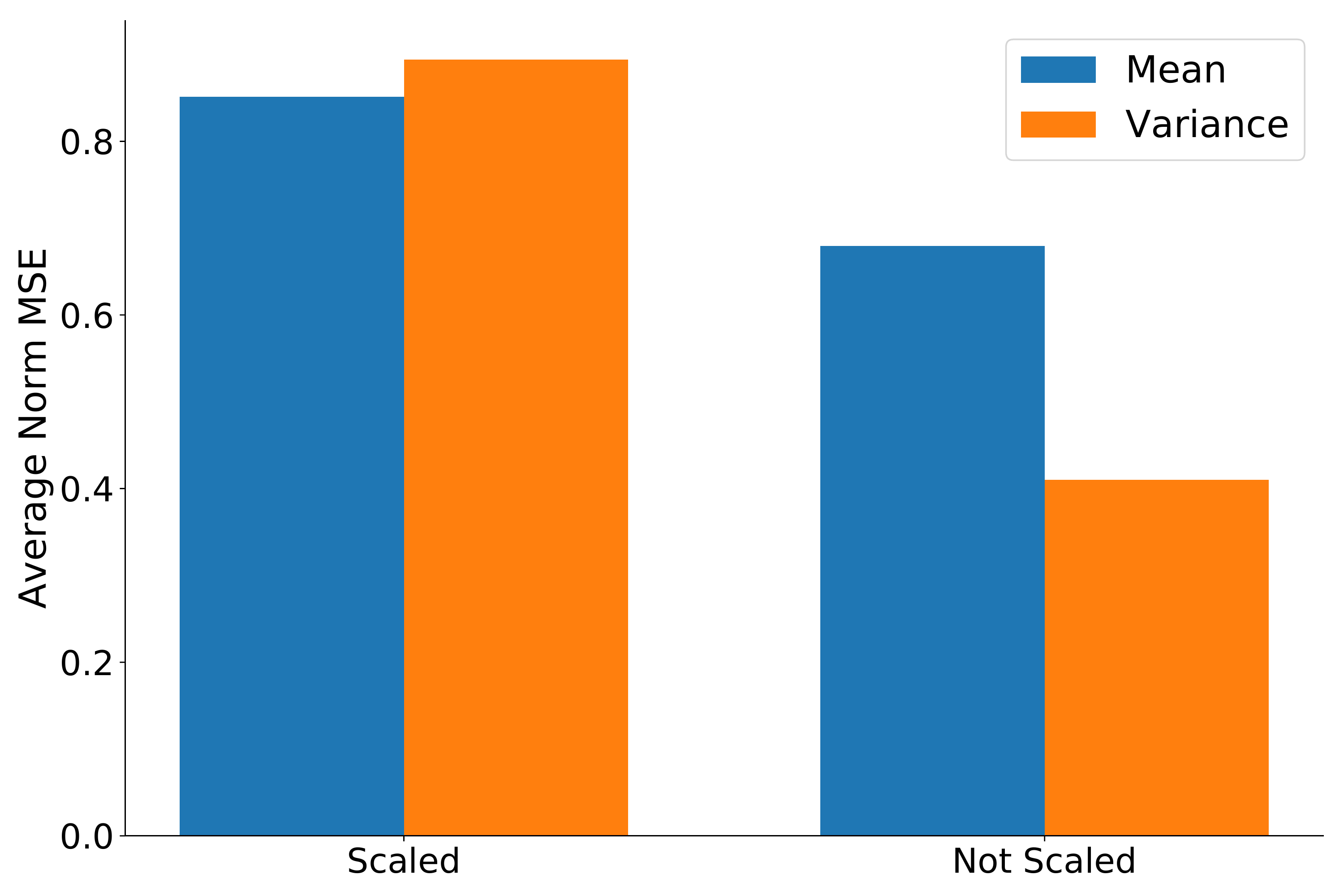}
			\caption{Average Norm MSE}
			\label{fig:sq:compare_scale_bar}
		\end{subfigure}    
		\caption{\textbf{Square Wave}. Each training run lasted for 50k timesteps and for each series 100 different runs were made. We show the normalized errors as a function of the prediction timescale, given on the $\tau$-scale. Results are averaged over the last 5k timesteps. For each $\tau$ we normalize by the maximum mean error across the series in the plot.  \textbf{Inputs)} Comparing the effect of using different timescale representations as input. As expected, providing timescale as $\gamma$ did better than $\tau$ on the short timescales, but worse on the longer timescales, although this cross over occurred at a much longer timescale than expected. Providing both $\gamma$ and $\tau$ did the best of all, producing the lowest errors across all probe timescales as well as providing the lowest variability. \textbf{Distribution)} Here we compare the effects of drawing $\kset$ from different distributions. As previously discussed, $\Kset_t$ was 6 elements long, always including $\tau\in\{1,100\}$, and sampling the additional 4 $\kset$. Excluding $\tau\in\{1,100\}$ we see that drawing all $\kset$ from the $\gamma$ scale performs better than drawing all from $\tau$ scale at shorter time scales, but does worse at longer timescales. Drawing half from each tends to follow the lower errors at all the timescales. \textbf{Scaling)} We compare the effects of scaling the cumulant. Here we see that scaling does improve performance on the shorter timescales, but causes worse performance on the longer ones.}
		\label{fig:square_exp}
	\end{figure*}
	
	\floatname{algorithm}{Algorithm}
	\begin{algorithm}[ht!]
		\caption{\ \ Generalization over $\gamma$ with TD(0)}\label{alg:general}
		\begin{algorithmic}
			\STATE \textbf{Input:} Feature representation $\x\in\reals^n$, policy $\pi$, and step-size $\alpha$
			\STATE \textbf{Output:} Vector $\w$.
			\STATE Initialize $\w\in\reals^n$ arbitrarily 
			\WHILE {$S'$ is not terminal}
			\STATE Observe state $S$, take action $A$ selected according to $\pi(S)$, and observe a next state $S'$ and cumulant $C$
			\STATE Pick a set of $\kset$ to train on:\\
			$\Kset\leftarrow \gamma SelectionFunction(Terminal=False)$
			\STATE $\boldsymbol{\Delta}\leftarrow \mathbf{0}$; Zeros vector, length $n$
			\FOR {$\kset$ in $\Kset$}
			\STATE $\delta = \Cumulant + \kset \x(S_{t+1},\kset)\tr\w - \x(S_t,\kset)\tr\w$\\
			$\boldsymbol{\Delta} \plusequals \delta\x(S_t,\kset)$
			\ENDFOR
			
			\STATE $\w = \w + \alpha \boldsymbol{\Delta}$
			
			\ENDWHILE
			
			\STATE $\Kset\leftarrow \gamma SelectionFunction(Terminal=True)$
			\FOR {$\kset$ in $\Kset$}
			\STATE $\delta = \Cumulant - \x(S_t,\kset)\tr\w$\\
			$\boldsymbol{\Delta} \plusequals \delta\x(S_t,\kset)$
			\ENDFOR
			
			\STATE $\w = \w + \alpha \boldsymbol{\Delta}$
		\end{algorithmic}
		\label{alg:linear_TD0}
	\end{algorithm}
	
	\subsection{Predictions on a Robot Arm}
	
%
%
	
	\begin{figure}[ht!]
		\centering
		
		\begin{subfigure}[t]{0.49\linewidth}
			\centering
			\includegraphics[height=2.0in]{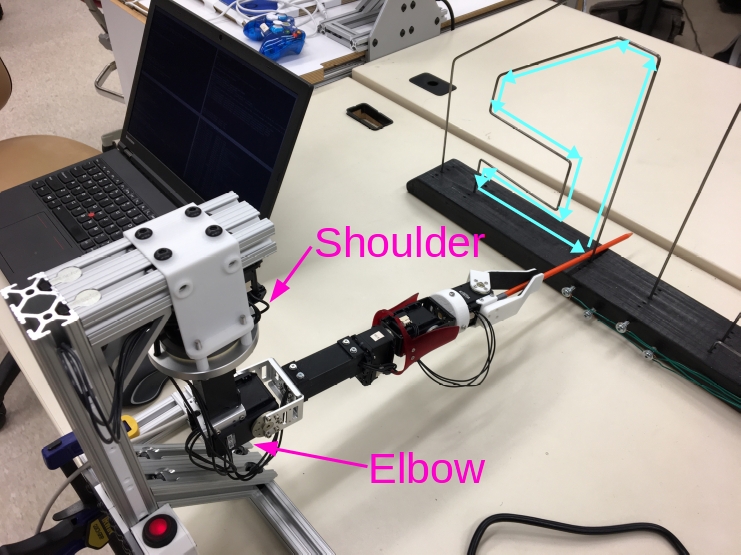}
			\caption{}
			\label{fig:robot_maze}
		\end{subfigure}
		\hfill
		\begin{subfigure}[t]{0.49\linewidth}
			\centering
			\includegraphics[width=0.95\textwidth]{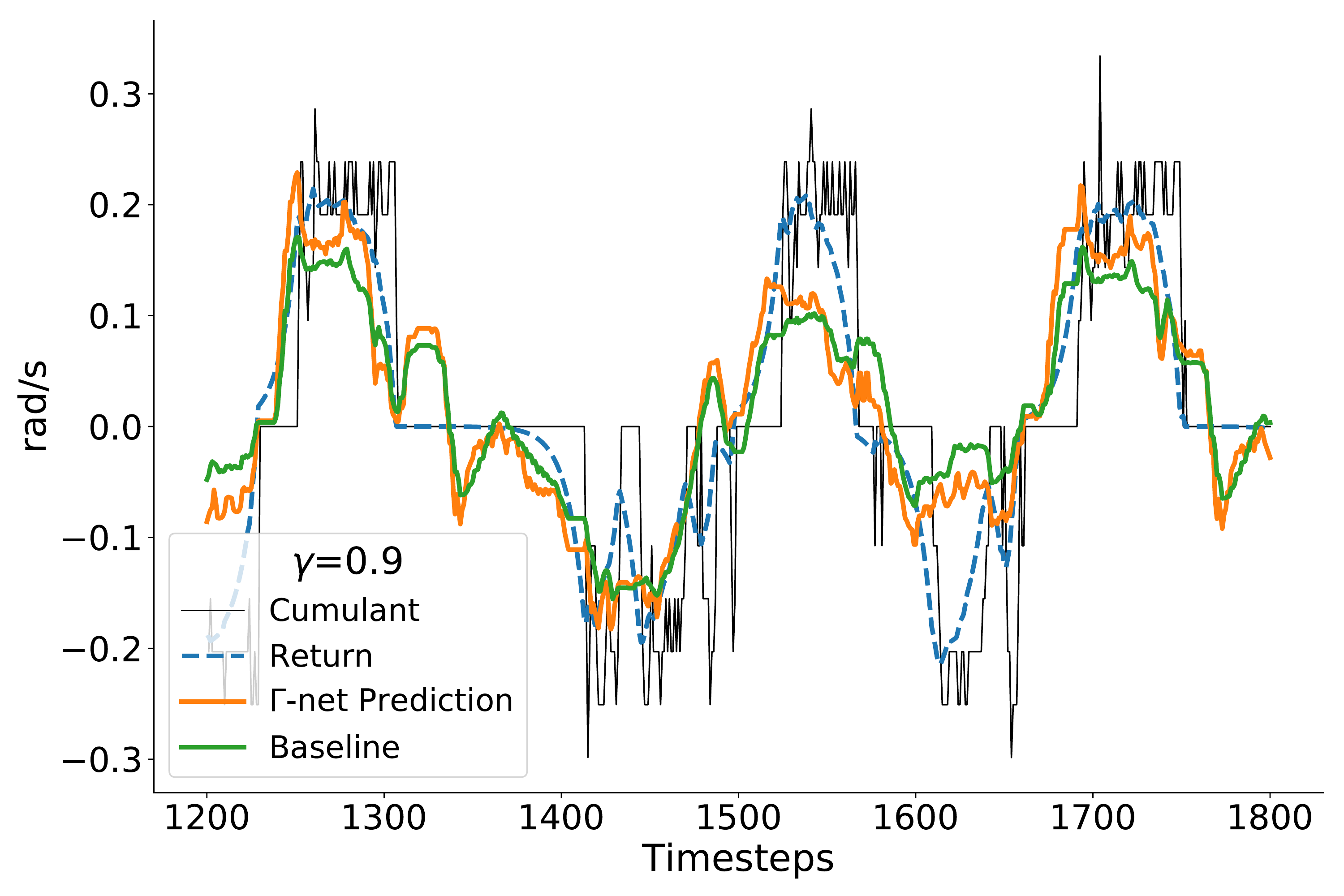}
			\caption{}
			\label{fig:robot_p_0.9}
		\end{subfigure}
		\\
		\begin{subfigure}[t]{0.49\linewidth}
			\centering
			\includegraphics[height=2.0in]{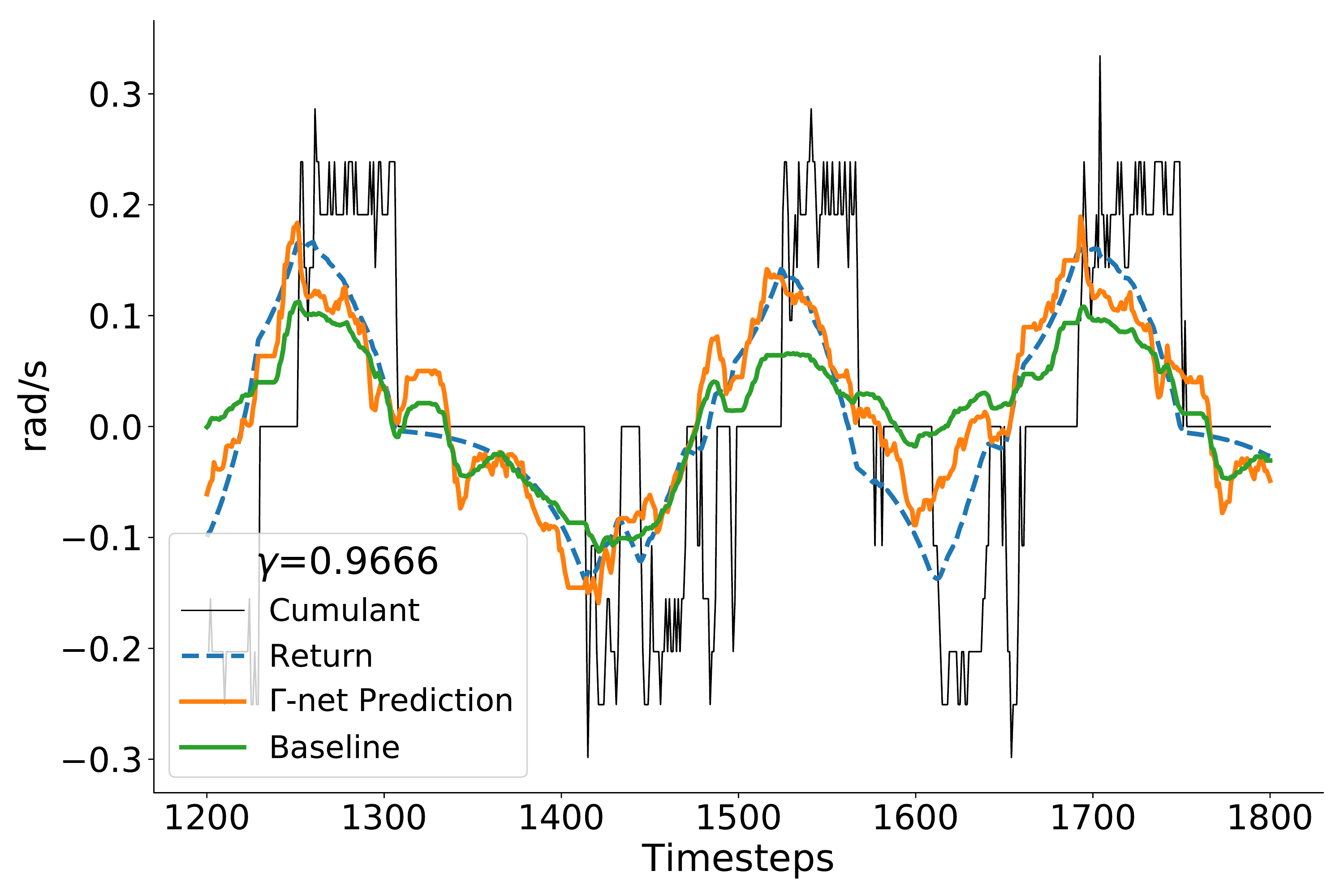}
			\caption{}
			\label{fig:robot_p_0.9666}
		\end{subfigure}
		\hfill
		\begin{subfigure}[t]{0.49\linewidth}
			\centering
			\includegraphics[width=0.95\textwidth]{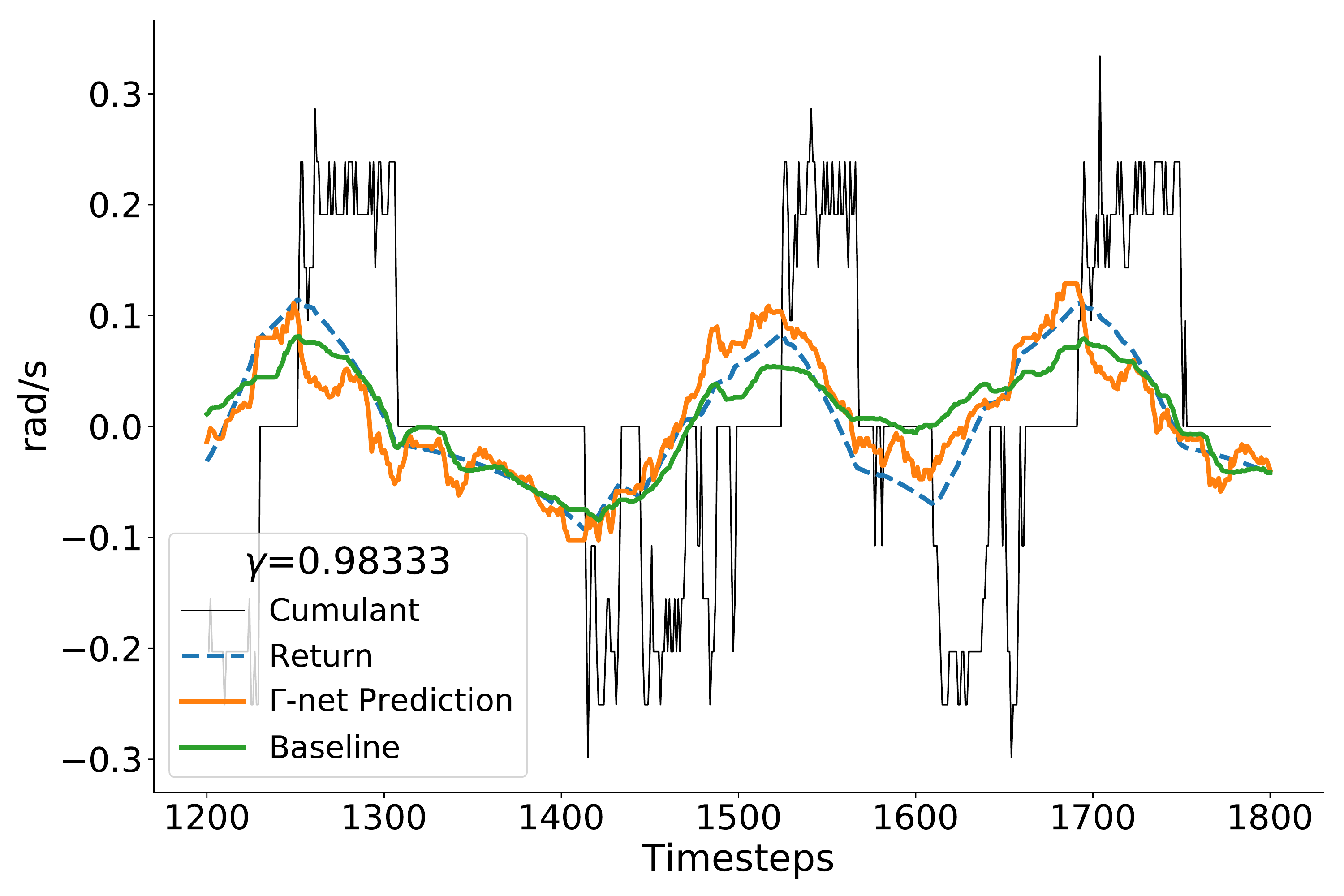}
			\caption{}
			\label{fig:robot_p_0.99}
		\end{subfigure}
		
		\caption{\textbf{Robot Arm.} \textbf{a}) A user controls the shoulder and elbow joints of a robot arm via joystick to move a rod counter-clockwise around a wire maze. \textbf{b-d}) Predictions of the speed of the shoulder joint. Predictions and returns are scaled by $(1-\gamma)$ for display purposes. Timescales correspond to $\sim$0.33,1.0, 2.0 s. {\thename} predictions (orange) perform similarly to the baseline (green) in matching the return (blue).}
		\label{fig:robot}
	\end{figure}
	
	In this experiment a human operated the shoulder rotation and elbow flexion joints of a robot arm by joystick. The task was to maintain contact between a rod held by the robot and the inside of a wire maze while moving in a counter-clockwise direction (Figure~\ref{fig:robot_maze}). Fifty circuits of the maze were completed in approximately 12 minutes. Network inputs were the normalized positions of the shoulder and elbow servos as well as both $\gamma$ and normalized $\tau$. Inputs were tilecoded \citep{Sutton1998} with 100 tilings of width 1.0 into a space of 2048 bits and a bias unit was added giving a feature vector of 2049 bits. Value estimates were computed by linear function approximation (LFA) and trained by TD(0). On each timestep $\Kset_t$ was generated from $\tau\in[1,100]$ ts. The upper and lower bounds were included in the set and one $\gamma$ and 29 $\tau$ were sampled uniformly from their respective scales for a total of 32 timescales. More emphasis was placed on sampling from the $\tau$ scale because of the relatively high update rate (30 Hz, 1 ts $\sim$ 0.03 ms). Thus, the likely important timescales will be above $\gamma=0.9$. Loss prescaling was used. We used a step-size of 0.1 divided by the number of active features. The step-size was linearly decayed to zero over the course of the training set. A baseline predictor with a fixed timescale was also trained using the same parameters as the {\thename} excepting the inclusion of timescale input.
	
	Figure~\ref{fig:robot} shows the {\thename} predicting shoulder joint speed at several timescales. Table~\ref{table:robot} shows the cumulative sum of absolute error between the predictor and the return over the whole dataset after training. With this configuration the {\thename} outperformed the baseline for most of the timescales tested. For $\gamma=0.99$ the baseline performed slightly better.
	
	\begin{table}[ht!]
		\caption{\textbf{Robot Arm.} Cumulative absolute error over the entire dataset after training. Lower is better.}
		\centering
		\begin{tabular}{|c|c|c|}
			\hline
			$\gamma$ & {\thename} & Baseline \\
			\hline
			0.9 & \textbf{1025} & 1124 \\
			0.9666 & \textbf{602} & 822 \\
			0.98333 & \textbf{379} & 440 \\
			0.99 & 273 & \textbf{253} \\
			\hline
		\end{tabular}
		\label{table:robot}
	\end{table}
	
	\subsection{Atari Environment}
	
	\newcommand{\comparisonplot}[2]{
		\begin{figure}[ht!]
			\centering
			
			\begin{subfigure}[t]{0.32\linewidth}
				\centering
				\includegraphics[width=0.98\textwidth]{img/p_e/#1/by_ts/MSE.pdf}
				\caption{MSE}
				\label{fig:#1_by_ts_MSE}
			\end{subfigure}
			\begin{subfigure}[t]{0.32\linewidth}
				\centering
				\includegraphics[width=0.98\textwidth]{img/p_e/#1/by_ts/MSE_bar.pdf}
				\caption{Avg Norm MSE}
				\label{fig:#1_by_ts_MSE_bar}
			\end{subfigure}
			\begin{subfigure}[t]{0.32\linewidth}
				\centering
				\includegraphics[width=0.98\textwidth]{img/p_e/#1/by_ts/corr.pdf}
				\caption{Correlation}
				\label{fig:#1_by_ts_corr}
			\end{subfigure}
			
			\caption{#2}
			\label{fig:#1}
		\end{figure}    
	}
	
	We examined the performance of {\thenames} under policy evaluation in the Arcade Learning Environment (ALE) \citep{Bellemare2015}. The agent's policy was trained using the Dopamine project's \citep{Dopamine} implementation of the Rainbow agent \citep{Hessel2018}, which uses the same network architecture as the DQN agent \citep{Mnih2015}, but adds prioritized replay \citep{Schaul2016}, n-step returns, and distributional representation of the value estimates \citep{Bellemare2017}. 
	
	The primary results presented are for the game Centipede with a Rainbow agent trained for 25 M frames, which we will refer to as \textit{Centipede@25M}. Additional Atari games were evaluated using agents trained for 200 M frames, which we will refer to as \textit{Atari@200M}. These agents were included as part of the Dopamine package and trained according to the specifications given in \cite{Dopamine}. Results for Atari@200M are included in the appendix.
	
	Figure~\ref{fig:centipede_predictions_in_time} shows predictions on the early transitions of a single episode. For this episode the expected return was estimated by running 2000 Monte Carlo rollouts from each state visited along the way (dashed lines). The solid lines indicate the {\thename} predictions after training for 20 M frames (using the \textit{direct} configuration which will be described in following sections).
	
	\begin{figure}[ht!]
		\begin{center}
			\centerline{\includegraphics[height=\imgheight]{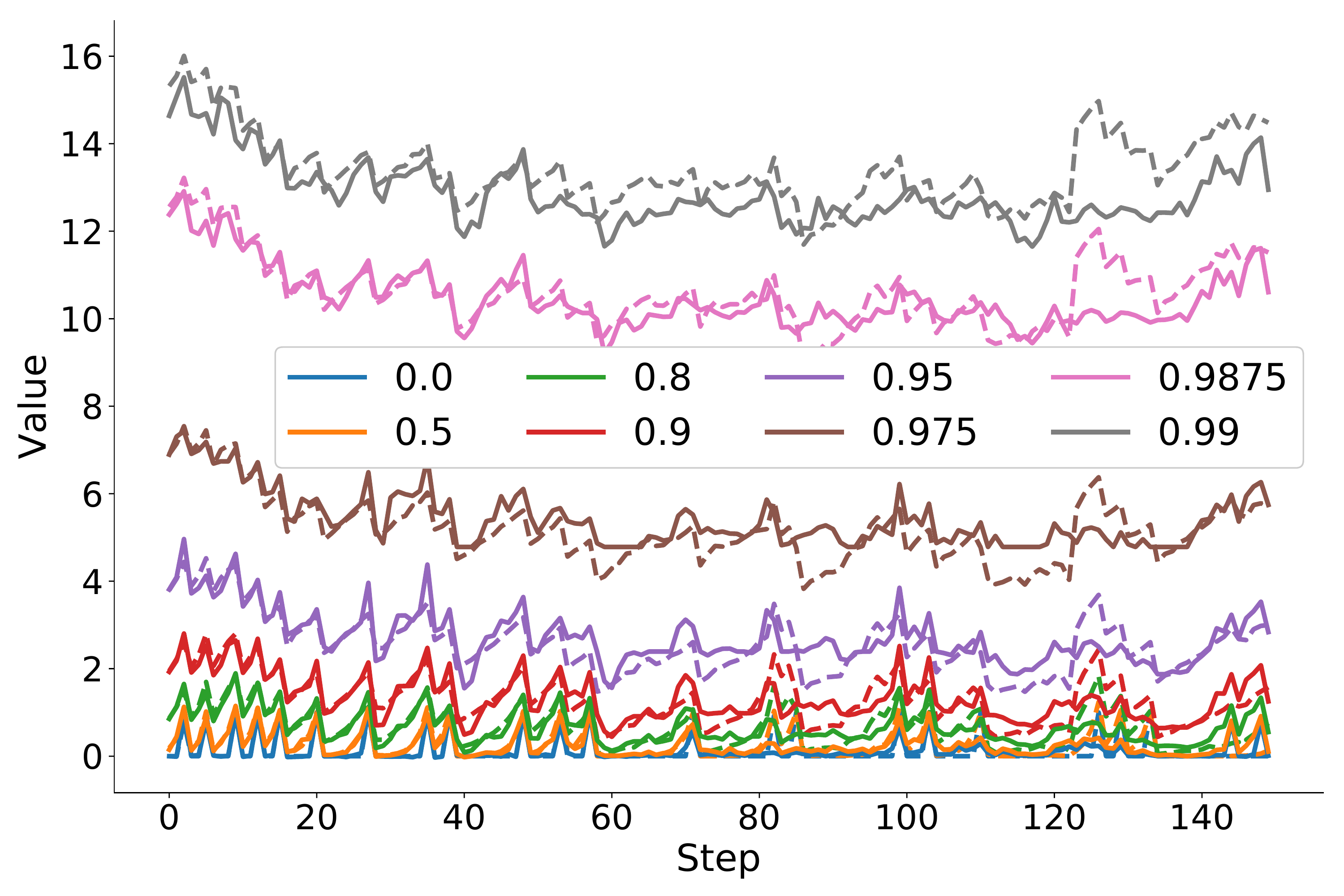}}    
			\caption{Predictions on Centipede@25M for different $\gamma$ from the start of a single episode. {\thename} predictions are shown in solid lines and the expected return, produced by Monte Carlo rollout, is shown by the dashed lines.}
			\label{fig:centipede_predictions_in_time}
		\end{center}
	\end{figure}
	
	\subsubsection{Training} 
\begin{figure}[ht!]
		\begin{center}
			\centerline{\includegraphics[width=0.8\linewidth]{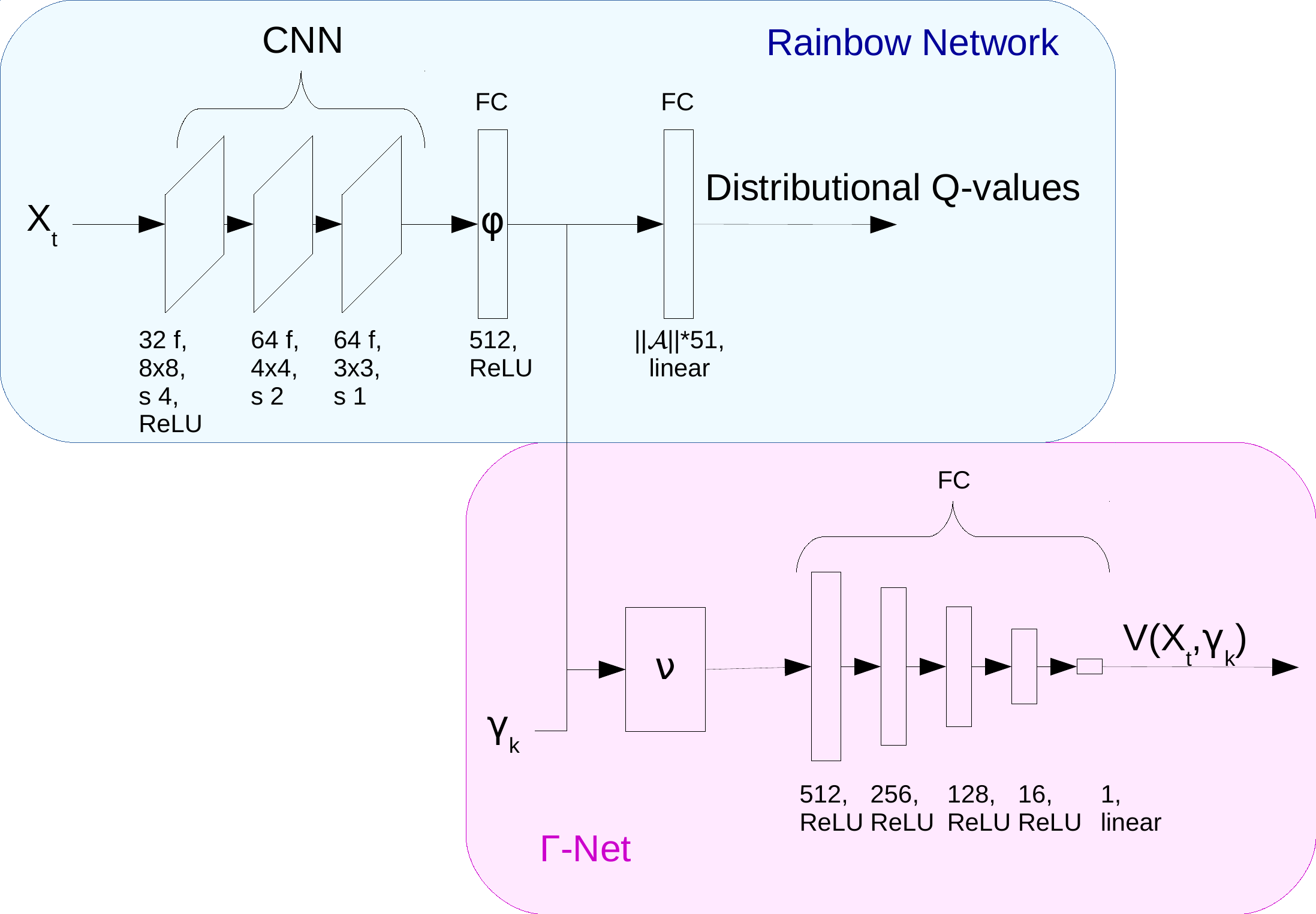}}
			\caption{Architecture used for the policy evaluation experiments. The Rainbow Network, from Dopamine, is used to generate episodes of data where we save the feature vector, $\phi$, to file. To train the {\thename} we then read in these files, store them in a prioritized replay buffer and sample from this replay buffer. The feature vector, $\phi$, is then combined with the $\kset$ using the embedding function $\nu$ which acts as input to the {\thename}.}
			\label{fig:architecture}
		\end{center}
	\end{figure}

The prediction networks were trained using samples of transitions generated by pretrained policies. Agents select actions using $\epsilon$-greedy over their Q-values. During policy training $\epsilon=0.001$, but for generating the samples used for training the {\thenames} we use an evaluation mode where $\epsilon=0.0001$. Transitions were generated sequentially and the environment was reset at the end of each episode or 27,000 steps, whichever came first. These transitions were saved to file in sequence and for each experiment they were reloaded in the same order. For each transition, we saved the reward as well as the activation of the final core layer of the agent's network $\x$, which serves as the input to the {\thenames}. The {\thename} network was composed of five fully-connected layers of sizes [512, 256, 128, 16, 1], with all but the final layer using ReLU activation. The architecture used in shown in Figure~\ref{fig:architecture}. Training of the {\thenames} proceeded as if the data was generated in an online fashion, as would be the case during policy learning. That is the agent would read in transition samples from the file, add them to a prioritized replay buffer, and then train by sampling from the replay buffer. When a new sample was added to the buffer it was given the highest level of priority so that its probability of being sampled was high. Like the policy training we train on a batch of sampled transitions, using n-step returns. To update the priorities for a given sample in the batch we use the maximum squared loss across $\Kset_t$.
	
	A $\Kset_t$ of size 8 was used, which always included lower and upper bounds of $\tau=[1,100]$. An additional 6 $\kset$ were drawn on each timestep. Unless otherwise stated the sampling was done by drawing 3 timescales uniformly each from the $\gamma$ scale on $[0, 0.99)$ and the $\tau$ scale on $[1, 100)$ (for $\tau$ we drew from the integer scales, rather than float). Each network was trained for 20 M frames with network weights saved every 500k frames. Additional training details can be found in the appendix. 
	
	\subsubsection{Evaluation} 
	To evaluate predictive accuracy we created a set of evaluation points for each game. These were generated by running the agent in evaluation mode over multiple episodes. At the start of each episode an offset was randomly chosen between $[10, 100)$ steps. Then, starting at the offset, the state of the environment and agent were saved every 30 steps (120 frames with 4 frame frameskip). For Centipede@25M a total of 269 evaluation points were created in this way including the episode start state. From each of these evaluation points we ran 1000 episodes till termination and then computed the average return.
	These were used as the baseline against which we computed our prediction error. To compute the prediction error for a given evaluation point we restored the agent's and environment's state and recorded the network's predictions for the probe timescales $[1, 2, 5, 10, 20, 40, 60, 80, 100]$ ts. For comparison, we trained fixed timescale networks for all the probe timescales (plotted in fuschia). These networks used exactly the same architecture as the \thename, but did not provide timescale as input to the network and only trained on the single fixed timescale. These probe networks also used loss scaling. For the \textit{Atari@200M} results a reduced set of probe timescales was used: $\tau=\{1, 10, 20, 40, 100\}$.
	
	We use a reference configuration of the {\thename} across the different plots. We plot this series in black and refer to it as \textit{direct} although the figure legends may give it a different label to call out the significance of its configuration for a given comparison. For this configuration both $\gamma$ and $\tau$ were provided as inputs to the network. Additionally, $\Kset_t$ was populated by drawing samples from both $\gamma$ and $\tau$ scales and loss scaling was used. For each of the other configurations only a single setting was modified from this reference.
	
	\subsubsection{Plotting}
	
	We focus our evaluation on the steady-state performance of the network, computing averages over the last 5 M frames of the 20 M frame runs (with evaluation at every 500k frames). Mean-squared error (MSE) for each experiment is presented as a function of the evaluation probe timescale given in $\tau$ (Ex. Figure~\ref{fig:fixed_vs_gamma_net_by_ts_MSE}). For each $\tau$ we normalize across the different series by the largest mean error. Thus, the largest mean error for each $\tau$ is shown as 1.0. We do this to be able to clearly show results for all the different timescales in a single plot despite the large differences in magnitude. As a result, series can only be directly compared within a plot, not across plots. To rank each for comparison we provide a bar chart (Ex. Figure~\ref{fig:fixed_vs_gamma_net_by_ts_MSE_bar}) which averages the normalized means and normalized variances of the MSE. That is, we take the normalized mean MSE for each $\tau$ and average across all $\tau$. Likewise we take the variance at each $\tau$, normalize it by the maximum variance for each $\tau$ and take the average across all $\tau$. Note that averaging this way is a biased approach in that it is dependent on what probe $\tau$ are used. For example, if we took many large $\tau$ and few small ones then our results would give more weight to the large $\tau$. In practice, the weighting of errors for different timescales will be task dependent.
	
	While conducting parameter sweeps it was observed that a particular network configuration might produce the lowest value of MSE but not actually be predictive. In this case the network would learn to always output a fixed value which captured the mean of the expected returns. Thus, we adopted a two step evaluation process. First, we took the evaluation points and concatenated them in sequence. We then computed the correlation between their expected returns and the predicted returns made by the network. If a configuration had a positive correlation then it would be considered for comparison with other architectures. We have also included the plots of correlation by probe timescale (Ex. Fig~\ref{fig:fixed_vs_gamma_net_by_ts_corr}). Correlation values are easily interpreted with the maximum (best) value of 1. This tells us how closely the shapes of the target sequence and the prediction sequence match.
	
	All series are an average over 6 seeds and the shading indicates max and min values. Note, that due to the high degree of overlap in many of the figures, color printing is required to discern individual series. Plots taken with respect to $\tau$ are produced by combining two different x-axes, allowing us to make both short and long timescales  discernible. This split occurs at $\tau=10$ and is indicated by the vertical black line.
	
	
	While our evaluation method seeks to discern differences in performance due to the various configuration, in reality most configurations perform similarly. In order to rank configurations we first considered the MSE and then variance.
	
	\subsubsection{Embedding Comparison.} We compare methods for combining the timescale inputs with the agent's features, $\x$, using an embedding vector $\e=\e(\x,\gamma)$ (Figure~\ref{fig:fixed_vs_gamma_net}). The \textbf{\textit{direct}} embedding performs a concatenation, $\e=[\x,\gamma]$. \citet{Xu2018} learned a vector, $\xi(\gamma)$, of size 16 which was concatenated with $\x$, which we refer to as \textbf{\textit{l\_embed}}. We also considered a Hadamard embedding in which a learned vector, $\xi(\gamma)$, the same length as $\x$, was combined using element-wise multiplication with $\x$, that is $\e=\x\odot\xi$ (\textbf{\textit{h\_l\_embed}}). Finally, we considered a matrix multiplication approach in which the timescales were given as inputs to a fully connected layer whose output was a square matrix, $\Xi(\gamma)$, with dimensions the same size as $\x$. The embedding was then formed by matrix multiplication: $\e=\x\tr\Xi$. We found little difference between the approaches in terms of their MSE or correlation. Overall the linear embedding appears the best choice based on its lower variance, but this did not hold universally for the other games evaluated (Figure~\ref{fig:200_games_embedding}). Learning and computation were both slower with the matrix multiplication approach (Figure~\ref{fig:embedding_2D}) and linear activations were generally slightly better than ReLU (Figure~\ref{fig:embedding_relu}).
	
	\comparisonplot{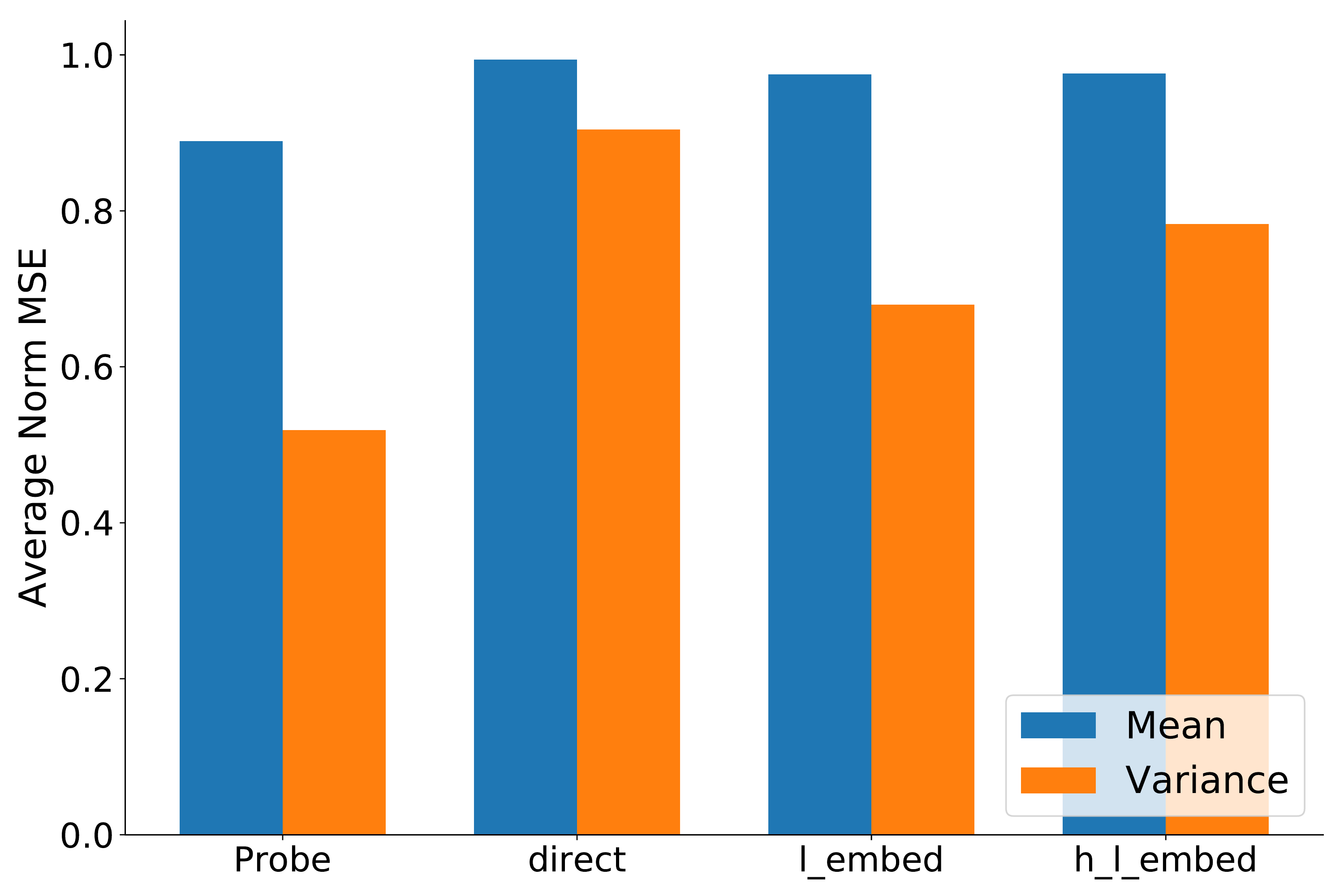}{\textbf{Embedding Comparison.} Several approaches for adding the timescale dependency to the network were investigated. \textbf{direct)} Concatenates the timescale with $\x$. \textbf{l\_embed)} Timescale is input to a fully connected layer of length 16 with linear activation whose output is concatenated with $\x$. \textbf{h\_l\_embed)} Timescale is input to a fully connected layer the same length as $\x$ with linear activation and then combined with $\x$ by element-wise multiplication. The \textit{l\_embed} approach appears to be slightly better due to its lower variance.}
	
	\subsubsection{Timescale Input Comparison} We examine how the input timescale representation affects prediction performance (Figures~\ref{fig:input_comparison}, \ref{fig:200_games_inputs}). We consider whether to use $\gamma$ or $\tau$ inputs or both. The $\gamma$ input values are naturally scaled between $[0, 1)$ and the $\tau$ input values were normalized by dividing by the max $\tau$, which in these experiments was 100. We consistently see that using only $\gamma$ produced the worst performance (Asteroids, in Figure~\ref{fig:200_games_inputs}, is an exception). Providing $\tau$ or both $\tau$ and $\gamma$ performed very similarly, but we consistently observed that providing both representations performed best for very short timescales and had lower variance.
	
	\comparisonplot{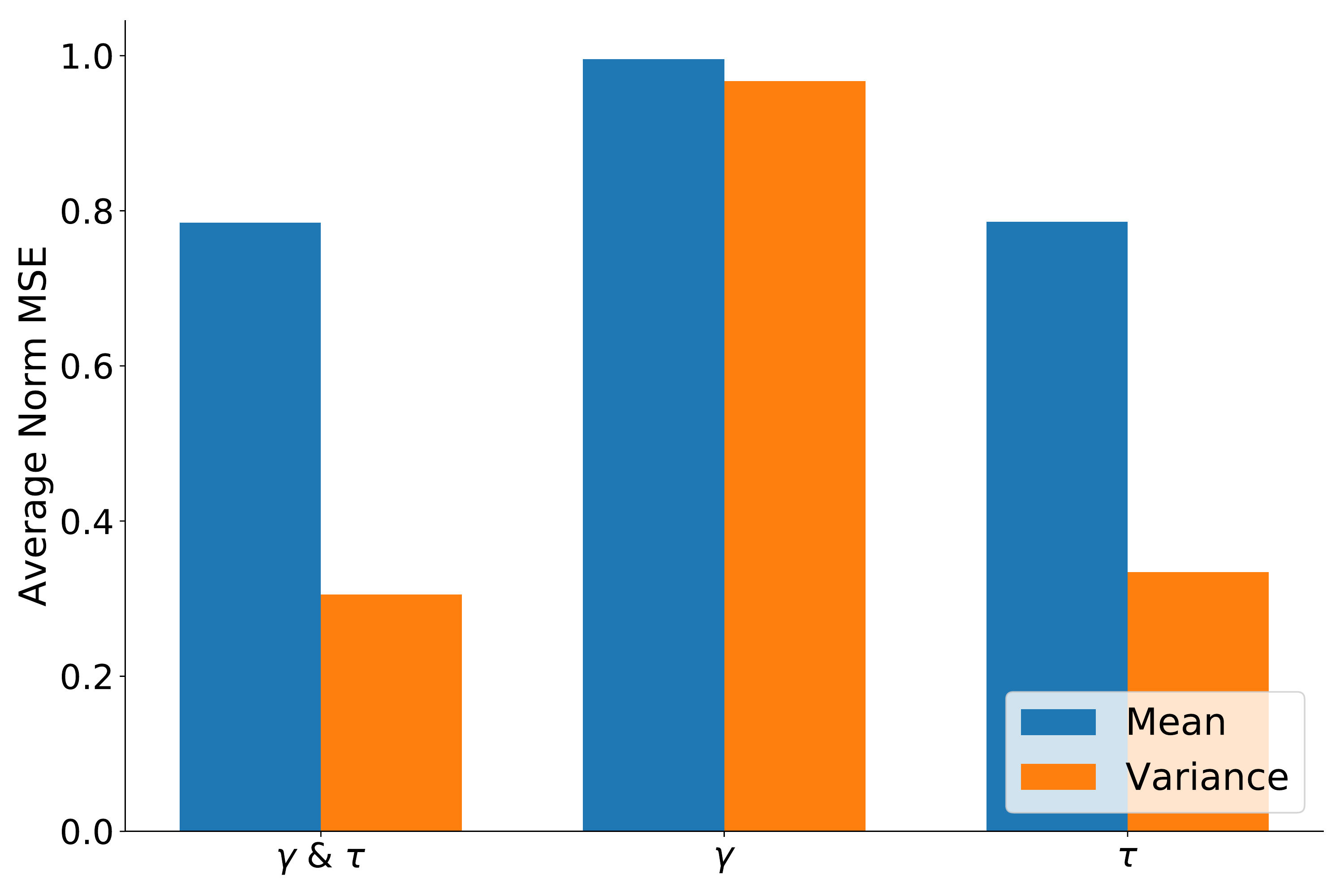}{\textbf{Input Comparison.} We compare performance of the network when providing $\gamma$, $\tau$ or both to the network inputs. We see that providing only $\gamma$ as the input timescale does the worst. Providing both $\gamma$ and $\tau$ or just $\tau$ perform similarly, but providing both does better at very short timescales.}
	
	\subsubsection{Distribution Comparison} We look at the effect of drawing $\Kset_t$ from different distributions (Figures~\ref{fig:dist}, \ref{fig:200_games_dist}). We use a $\Kset_t$ of size 8, two of which are always the lower and upper bounds $\tau=[1,100]$. Six additional $\kset$ are drawn from a given distribution. We either draw all six uniformly from the $\gamma$ or $\tau$ scale or draw half from each. We see that drawing solely from $\gamma$ performs worst overall, particularly at longer timescales, as is expected. Surprisingly, $\gamma$ did not consistently outperform $\tau$ at very short timescales. If we consider all timescales and games evaluated there is no clear winner between drawing solely from $\tau$ or from $\tau$ and $\gamma$. However, at very short timescales drawing from both tended to produce better results. Thus, we recommend drawing from both scales as a default.
	
	\comparisonplot{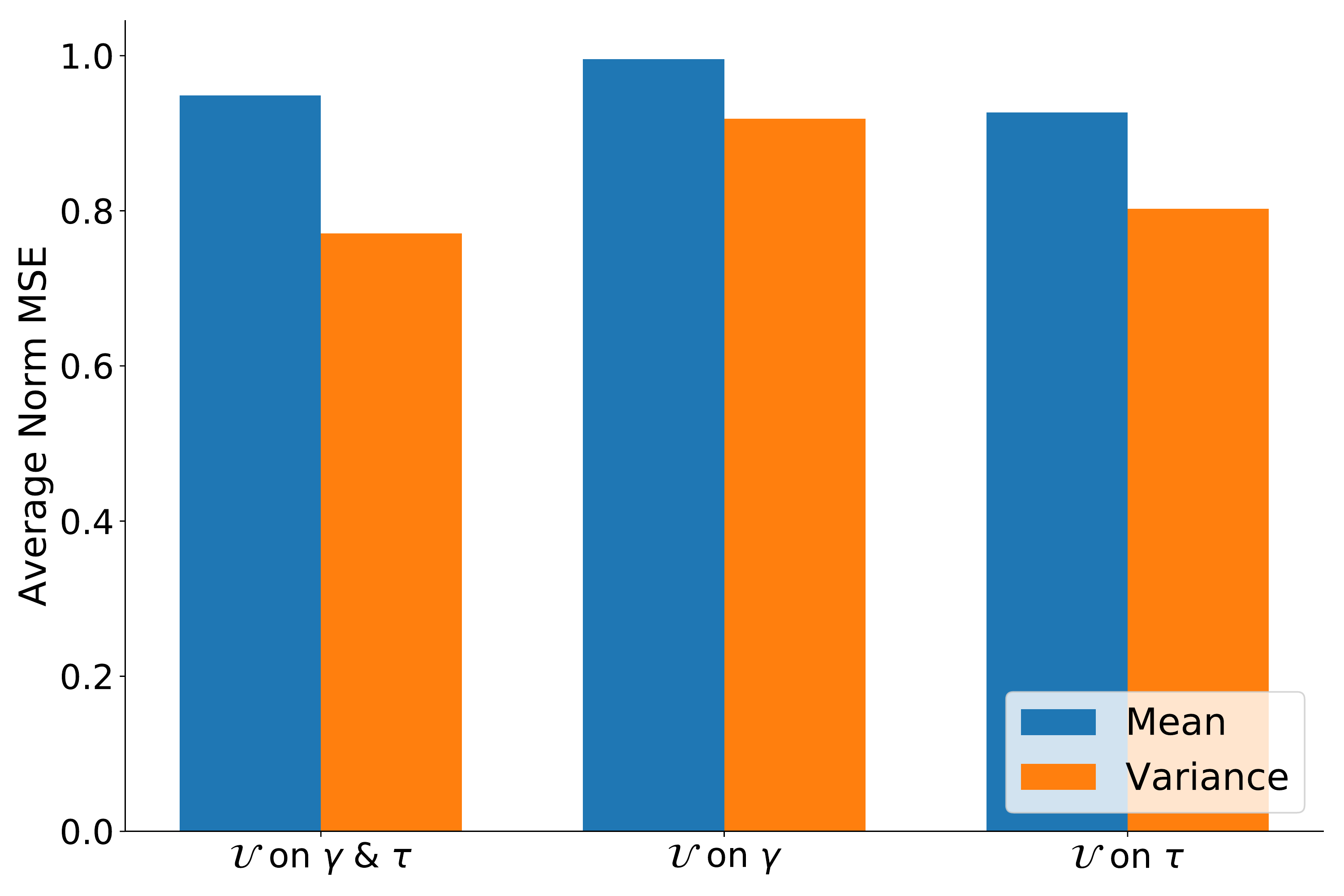}{\textbf{Distribution Comparison.} We compare different distributions used to generate $\Kset_t$. At lower timescales sampling from the $\gamma$ scale does better than sampling from the $\tau$ scale and the opposite holds at longer timescales. Sampling from both provides a compromise in performance.}

	\subsubsection{Loss Scaling} We examined the effect that loss scaling has on network performance. Figure~\ref{fig:scaling_comparison} shows that on Centipede@25M there is a clear benefit, with clearly lower MSE and variance. Scaling the loss was expected to improve short timescale performance. Surprisingly, in terms of MSE, the greatest impact was on longer timescales.  However, such a pronounced difference was not seen in other Atari games (Figure~\ref{fig:200_games_scaling}). Instead we saw a general trend in which scaling did improve performance at short scales at the cost of performance at mid and long timescales, which was in line with our expectations (again, Asteroids was somewhat an exception).
	
	\comparisonplot{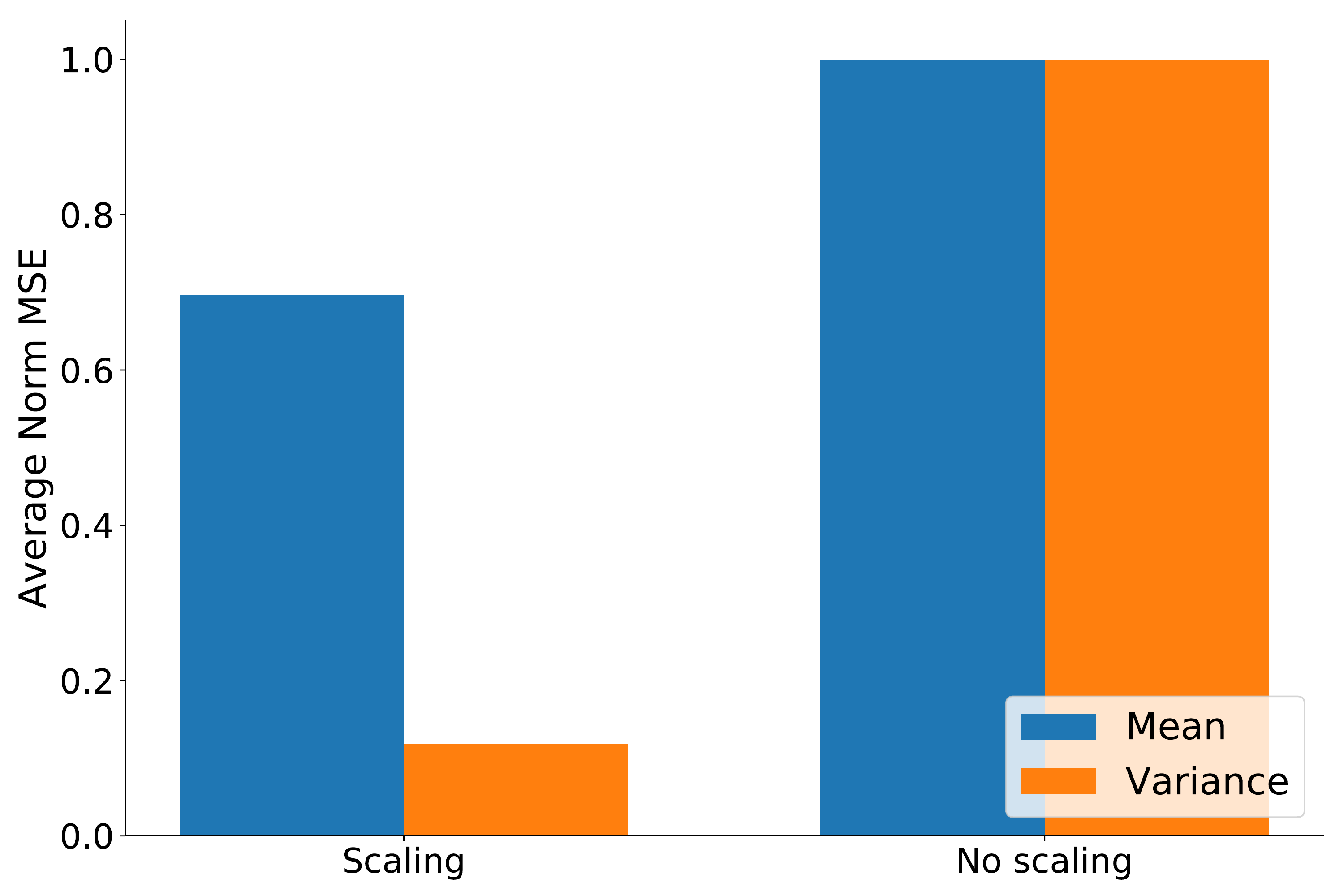}{\textbf{Scaling Comparison.} We examined the effects that scaling the loss by $(1-\kset)$ has. We see that scaling results in lower overall error and variance. Note that such a clear separation was not observed over other games tested.}
	
	\subsubsection{Estimation by Interpolation} An alternative approach to estimating value at arbitrary timescales is to have multiple prediction heads, each at a fixed timescale, and then linearly interpolate between the nearest bracketing timescales. In Figure~\ref{fig:interpolated} we show results for such an interpolation. Here we took the previously trained probe networks (with scaled loss and the taper network architecture) and performed linear interpolation for $\tau=[1.5, 3.5, 7.5, 15, 30, 50, 70, 90]$. Because of the non-linear relationship between $\tau$ and $\gamma$ the linear interpolation gives different weighting depending on whether the interpolation is done on the $\tau$ or $\gamma$ scale. Interpolating in these spaces is also compared. Results show that performance was fairly similar between the interpolation scales, but that the {\thename} did not perform as well. While it might have been expected that the ability of the neural network used by {\thename} to capture the non-linearity of the timescales would give it an advantage, this was not shown in this experiment. Rather, we suspect that the increased accuracy of the probe networks allowed the interpolation approach to win out.
	
	\comparisonplot{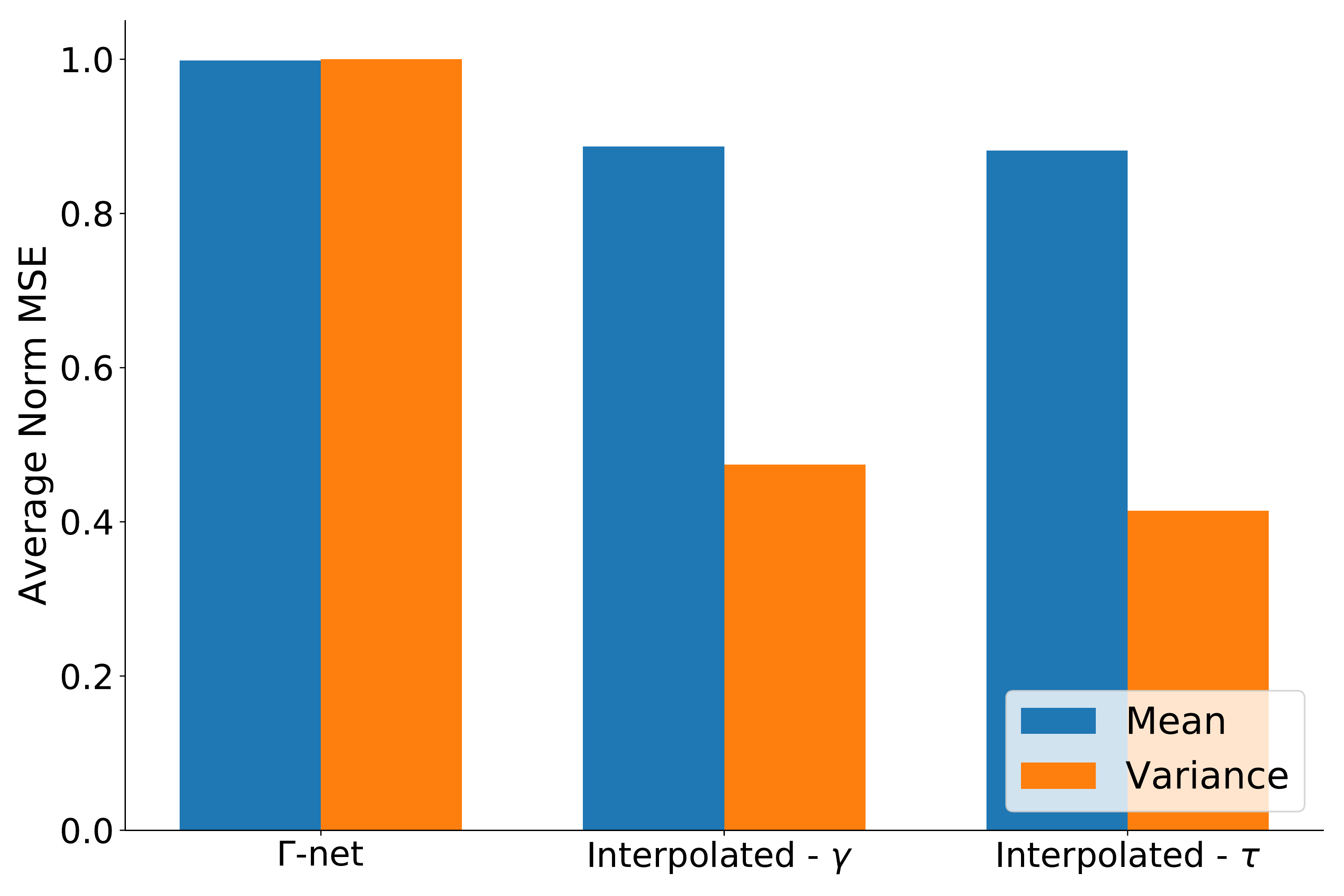}{\textbf{Interpolated.} Predictions are made between the probe timescales by taking the weighted average of the predictions made by the bracketing probe networks. Due to the non-linear relationships of $\gamma$ and $\tau$ different weightings are produced if the weighting is done in either scale. We see that either interpolation produces better results than the {\thenames}.}

\section{Discussion}

We have empirically evaluated various approaches to constructing {\thenames} and compared their predictive accuracy to baseline predictors. While we sought to separate the impacts of the various approaches, in reality all of the variants we explored performed similarly.  We have considered several different Atari games with deep learning architectures as well as a simulation signal and robotics demonstration using a shallow architecture. Overall we found that {\thenames} worked reliably both for reward and sensorimotor prediction. 

Despite the relatively minor differences in performance across the variants we do make some recommendations for implementation. Since there was no universal difference between the \textit{direct} or \textit{l\_embed} embedding approaches we recommend just using the simplest, \textit{direct}. If looking for a general approach that is not specifically adapted to the task then we recommend using both $\gamma$ and $\tau$ as inputs to the network as well as drawing samples from both scales in order to populate $\Kset_t$. On the other hand if longer timescales are preferred then it seems sufficient to use only $\tau$ for both input and sampling distribution. With regards to scaling the loss a clear universal benefit has not been observed and we suggest that further investigation is required to determine the best way to balance the losses resulting from different timescales. Such an investigation is a clear opportunity for future work. 

Our method is thus far limited to the fixed discounting case. However, one of the key generalizations of GVFs is to support transition-dependent discounting functions: $\gamma_{t+1}\equiv\gamma(S_t,A_t,S_{t+1})$ \citep{White2017}. This allows GVFs to be more expressive in terms of what the types of returns they can estimate. Extending our method to support such discounting is clearly an important next-step in this work. 

\begin{figure}[ht!]
	\begin{center}
		\centerline{\includegraphics[height=\imgheight]{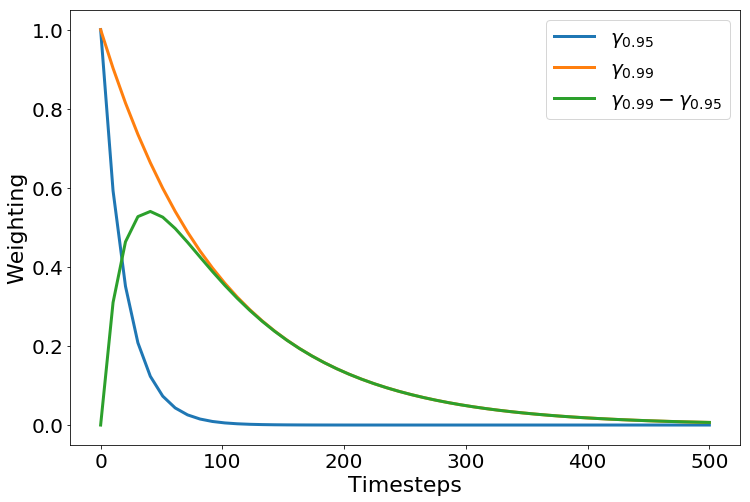}}
		\caption{Composing non-geometric returns (green) by taking the difference of two predictions at different timescales.}
		\label{fig:diff_of_gammas}
	\end{center}
\end{figure}

There are several ways in which our work and that of \citet{Fedus2019} are complementary. First, they demonstrated that using value predictions at many different timescales could serve as useful auxiliary tasks for driving representation learning. A clear next step is to investigate whether or not {\thenames} could also serve as a useful auxiliary task. One of the advantages of TD algorithms is that they allow the agent to bootstrap estimation of the return from its existing estimates. This limits a single predictor to only capturing returns with geometric discounting. However, such returns can be used as a basis to form alternative returns as is demonstrated by Figure~\ref{fig:diff_of_gammas}. In fact, \citet{Fedus2019} used geometrically discounted value estimates at multiple scales as a basis to estimate hyperbolically discounted returns. {\thenames} could provide such a basis function using a single network.

Long timescale predictions can be difficult to learn due to the higher variance of the returns. \citet{Romoff2019} presented an algorithm which computes values for an ordered set of timescales by predicting the differences between the values using separate network heads. Value estimates are constructed in a cascade where each timescale prediction adds to the one that came before it. They showed their method could improve estimation accuracy for longer timescales by leveraging the accuracy of the easier to learn shorter timescales. We might expect a similar effect using {\thenames} where long timescale predictions could benefit from the short timescales being learned directly in the network. Our current evaluation approach is not fine grained enough to discern such a benefit. Thus, this area warrants further exploration.

{\thenames} is related to other works which seek to learn many different predictions simultaneously and tractably. The UVFA \citep{Schaul2015}, on which this work is based, generalizes over goals. The successor representation (SR) \citep{Dayan1993} separates environment dynamics from reward, providing a way to transfer learning across tasks \citep{Barreto2017,Sherstan2018SR}. These ideas have been combined \citep{Mankowitz2018,Ma2018} to enable transfer learning over multiple goals using off-policy learning. However, these methods still use fixed timescales, thus, a natural extension of {\thenames} is to combine them with these approaches.

The original motivation for this work was to use {\thenames} to create GVFs which form a predictive representations of state for use by the agent's policy. It now seems that the best approach would be to use multiple heads with predictions at fixed timescales and let the policy network learn to generalize over those predictions as it needed. Such an approach could be costly in terms of network weights and {\thenames} might accomplish the same thing with a smaller network.
	
	\section{Conclusion}
	
	We presented {\thenames}, a simple technique for generalizing value estimation across timescale. This technique allows a system to make predictions for values of any timescale within the training regime of the network. We expect that this ability will be useful in areas such as predictive representations of state---i.e., modeling the world as a collection of predictions about future sensorimotor signals. In complex environments complete models are not feasible, thus, being able to query for predicted outcomes at any timescale makes a model potentially more compact and expressive. An investigation of {\thenames} in different control learning scenarios is an important area for future work, and we believe they may be of benefit to ongoing research in planning and lifelong learning. In particular {\thenames} are complimentary to approaches which seek to learn many things about the world simultaneously such as the successor representation and universal value functions, suggesting that {\thenames} may provide us with a functional new tool for the pursuit of knowledgeable intelligent systems.

\section*{Acknowledgements}

The authors would like to thank the following colleagues for providing thoughtful suggestions to this work: Alex Kearney, Marlos C. Machado, and Matt Schlegel. Additionally, Brendan Bennett, Jesse Farebrother and Vivek Veeriah provided helpful technical assistance. Initial stages of this work were funded by Cogitai and additional support was provided by the Natural Sciences and Engineering Research Council of Canada (NSERC), Compute Canada, the Canada Research Chair’s program, Alberta Innovates, and the Alberta Machine Intelligence Institute (Amii).

	\bibliographystyle{apa}
	\bibliography{main}
	
	
	\appendix
	\appendixpage
	
	\section{Atari Details}
	
	Various parameters are indicated in Table~\ref{tab:parameters}.
	
	A brief sweep was made over the step-size parameter (also referred to as \textit{learning rate}) for the Centipede@25M policy. Sweeps were made over the probe timescales as well as over various variants of the {\thename} for 3 seeds each. The values tried were: $6.25e^{-4},6.25e^{-5},6.25e^{-6}$. It was found that, almost universally, the value $6.25e^{-5}$ gave the lowest error when errors were aggregated over all probe timescales. This is also the step-size used in training the Rainbow agent. This step-size value was used for all reported experiments. Note that these sweeps were done on Centipede@25M experiments only.
	
	Dopamine's implementation of the prioritized replay buffer used fixed discounting for a single timescale. Thus, we needed to modify this implementation to return the n-step transitions and then apply discounting afterwards.
	
	We use a frame skip of 4, meaning that when an action is sent to the environment it is executed 4 times in a row and the resulting final frame is returned as observation. The implementation also uses frame stacking in which a max pooling is taken over the last 2 consecutive frames in order to deal with flicker in the rendering of the game images. We used sticky-actions with a probability of 0.25. This means that when an action is sent to the environment there is a 25\% chance that the environment will use the previous action instead. Every reset of the ALE environment restores the environment to the same initial state. State transitions are deterministic. The policy was trained with an $\epsilon$-greedy value of $0.001$, but for evaluation transitions were generated with $\epsilon$ reduced to $0.0001$. Thus, during evaluation the largest source of stochasticity is due to the sticky-actions. Further, the agent sees the early states of the episode more frequently than the later states. Like the policy training, one training update was performed for every 4 steps in the environment. Since every step in the environment corresponds to 4 frames a training update was performed every 16 frames.
	
	To train the sampled batch of transitions on $\Kset$ we tile the samples with each $\kset$. Thus, for a batch size of 32 sampled transitions and a $\Kset$ of 8 the effective batch size is 256. This does add some additional computation time to the process, but this is also affected by the quality of the implementation. When sampling from the $\tau$ scale for $\Kset_t$ we used the integer scale rather than float.
	
	Like the policy we use a target network which periodically copies weights from the online network; it is the online network which is updated on each training step and the target network which is used for bootstrapping. Note TD learning is typically trained using a semi-gradient approach in which the gradients are not computed with respect to the bootstrapping.
	
	\begin{table*}[!t]
		\centering
		\begin{tabular}{|l|c|}
			\hline
			Parameter & Value \\
			\hline
			Input dim & 84x84 \\
			$\phi$ dim & 512 \\
			Replay buffer size & 100000 \\
			Batch size & 32 \\
			n-step & 4 \\
			Min-replay history & 20000 \\
			Sync interval & 10000 \\
			Frameskip & 4 \\
			Sticky-actions & 0.25 \\
			Terminal on life loss & False \\
			Max steps per episode & 27000 \\
			Consecutive frame pooling & True \\
			$\epsilon$-greedy: policy learning & 0.001 \\
			$\epsilon$-greedy: evaluation & 0.0001 \\
			Adam optimizer: Step-size (learning rate) & $6.25e^{-5}$ \\
			Adam optimizer: eps & $1e^{-8}$ \\
			\hline
		\end{tabular}
		\caption{Parameters}
		\label{tab:parameters}
	\end{table*}
	
	\section{Additional Centipede@25M Figures}
	\label{sec:AdditionalAtariFigs}
	
\begin{figure*}[t!]
		\centering
		
		\begin{subfigure}[t]{0.49\linewidth}
			\centering
			\includegraphics[width=0.8\textwidth]{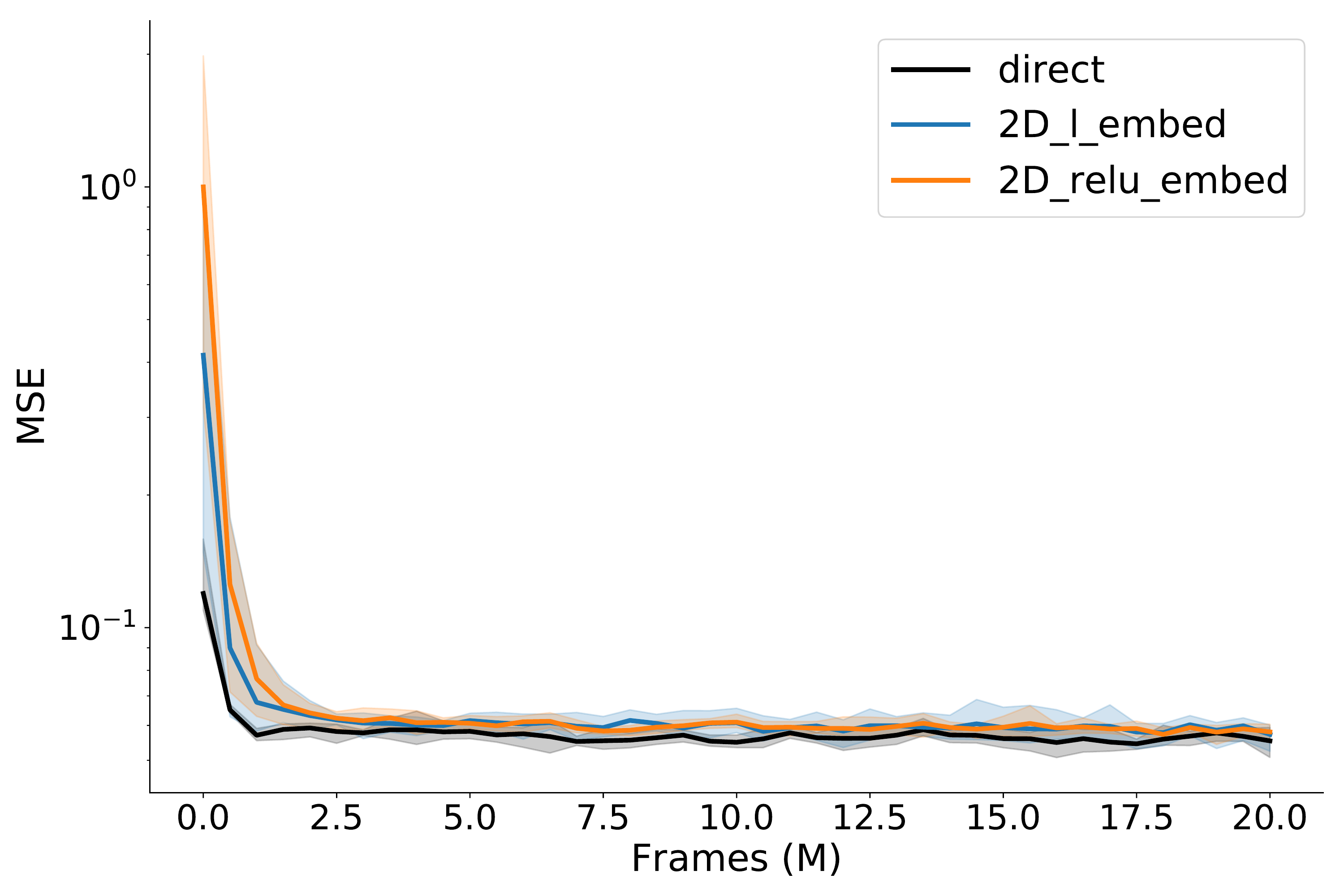}
			\caption{}
			\label{fig:embedding_2D_MSE}
		\end{subfigure}
		\hfill
		\begin{subfigure}[t]{0.49\linewidth}
			\centering
			\includegraphics[width=0.8\textwidth]{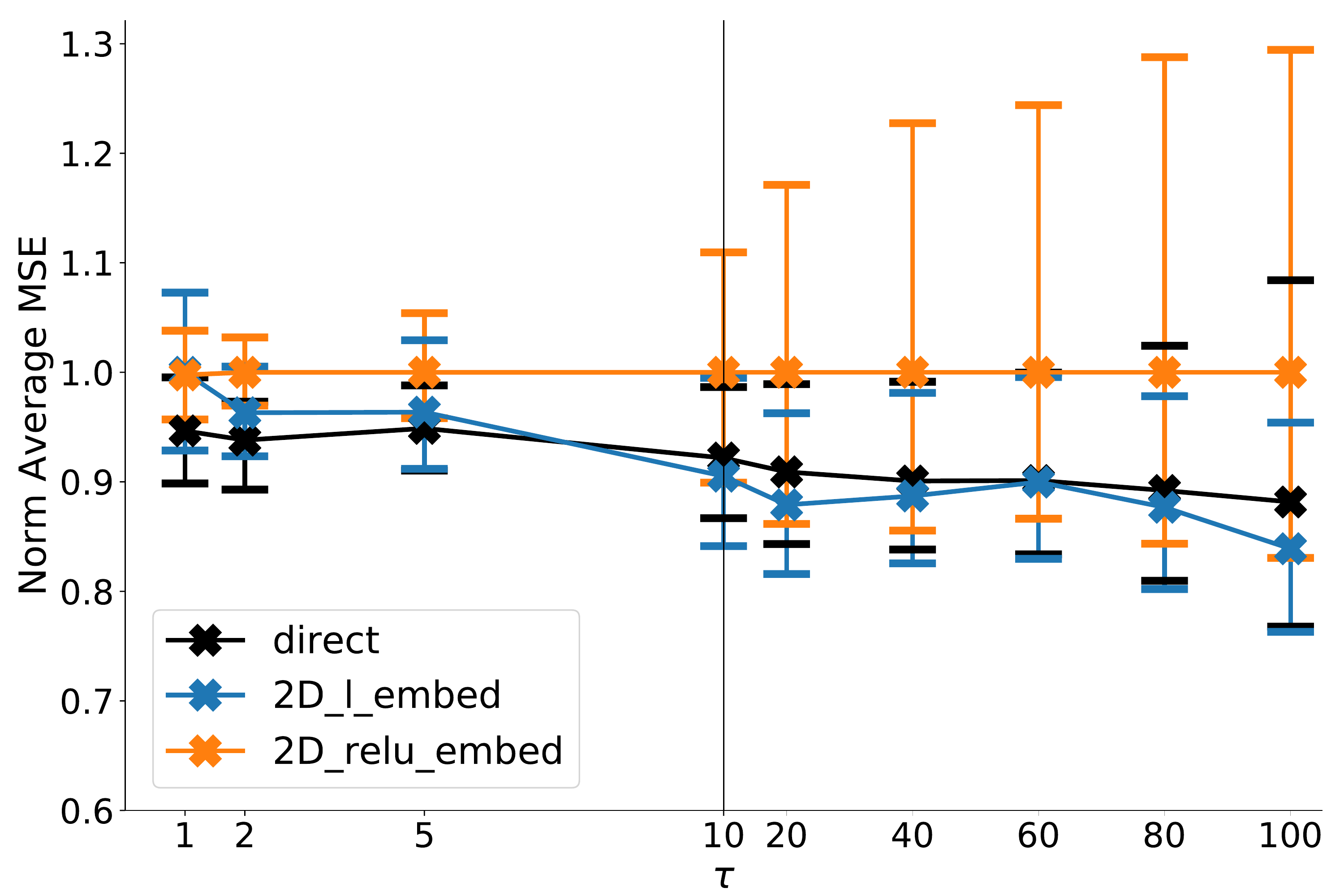}
			\caption{}
			\label{fig:embedding_2D_by_ts_MSE}
		\end{subfigure}
		\\
		\begin{subfigure}[t]{0.49\linewidth}
			\centering
			\includegraphics[width=0.8\textwidth]{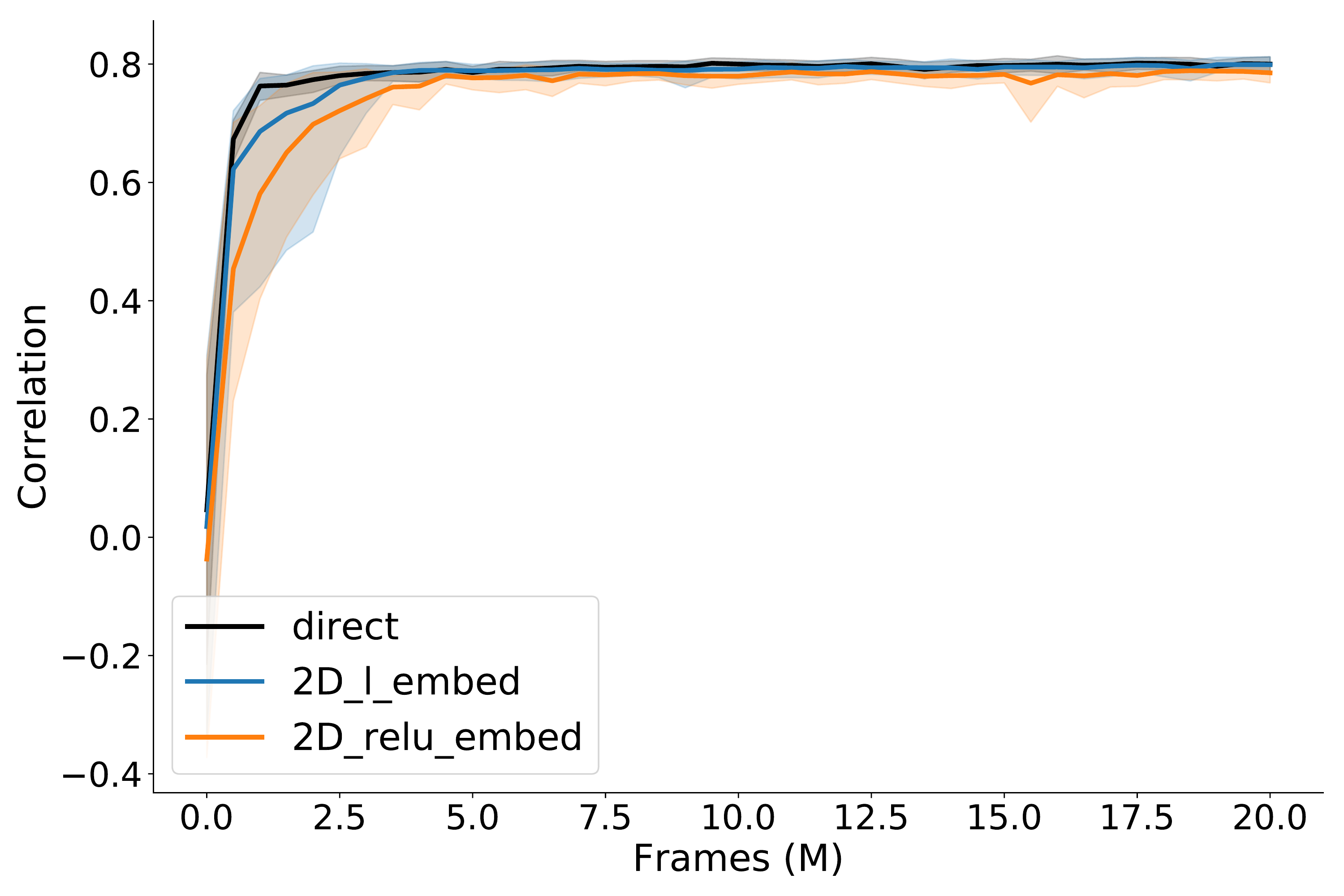}
			\caption{}
			\label{fig:embedding_2D_corr}
		\end{subfigure}
		\hfill
		\begin{subfigure}[t]{0.49\linewidth}
			\centering
			\includegraphics[width=0.8\textwidth]{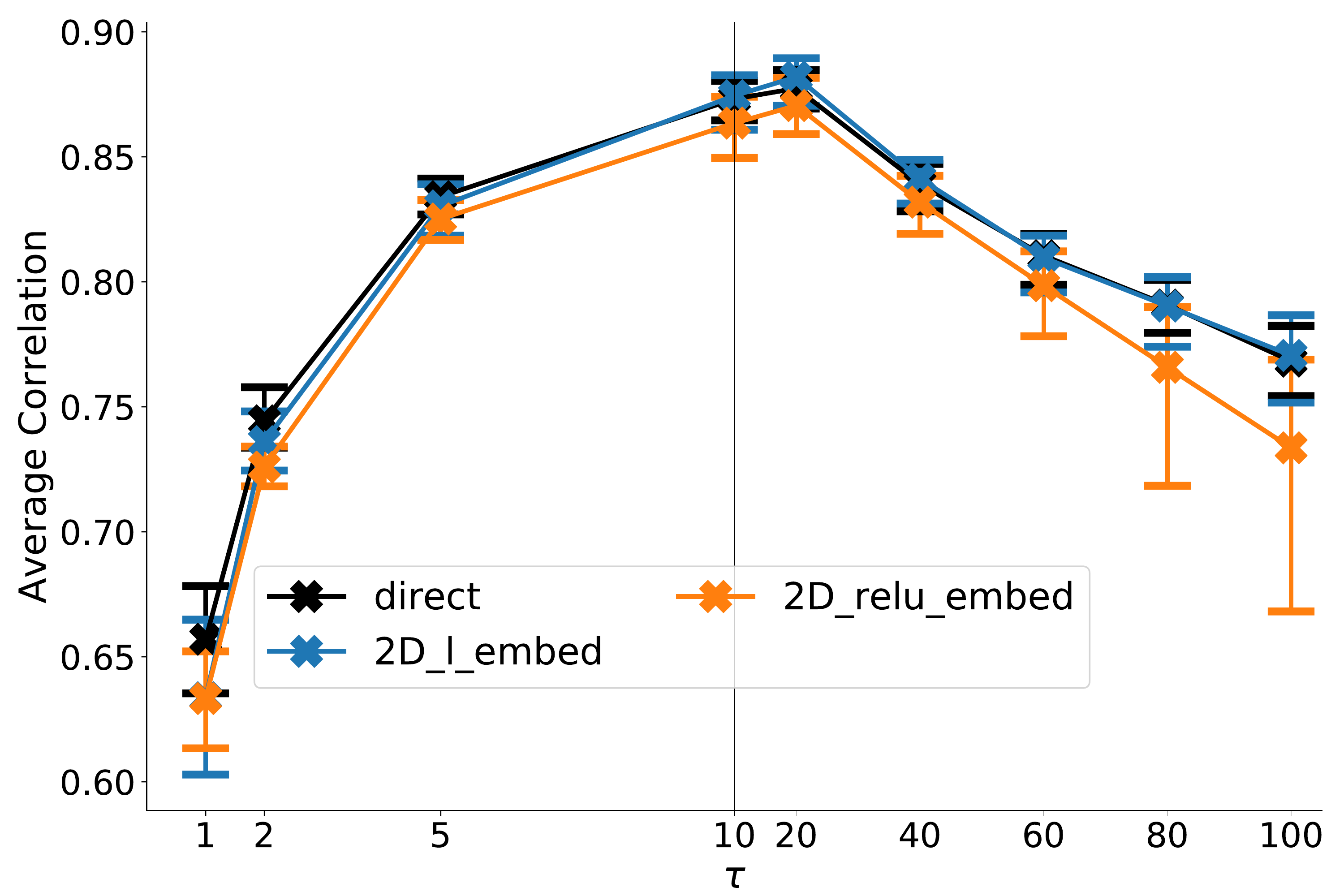}
			\caption{}
			\label{fig:embedding_2D_by_ts_corr}
		\end{subfigure}
		
		\caption{\textbf{Matrix Embedding.} Timescale embedding was performed by matrix multiplication, that is $\e_{t;k}=\x_t\tr\Xi_k(\kset)$. Where the size of $\Xi$ was chosen such that the multiplication produced an output vector the same size as $\x$. That is, $||\x||=512$ and $||\Xi||=512\times512$. The series \textit{2D\_l\_embed} and \textit{2D\_relu\_embed} use the matrix multiplication approach with either a linear or ReLU activation function.The \textit{direct} model is as described in the main paper. The matrix multiplication approach tends to learn much slower than the direct, with markedly reduced performance at the higher timescales. Further, consistent with the other embedding approach, ReLU activation performs worse than a linear one. Additionally, training the 2D models was computationally expensive and training was noticeably longer.}
		\label{fig:embedding_2D}
	\end{figure*}

	Here we present several additional figures of evaluation on the Centipede@25M policy. Figure~\ref{fig:embedding_relu} compares the performance of using linear or ReLU embeddings on the embedding networks used in Figure~\ref{fig:fixed_vs_gamma_net}. The ReLU embedding performs the same as the linear embedding for the concatenated architecture and performs worse with the Hadamard. Figure~\ref{fig:embedding_2D} looks at a matrix embedding approach. We see that here too the ReLU embedding performs worst. Note that with the matrix embedding the learning was slower than for the \textit{direct} embedding.
	
	\comparisonplot{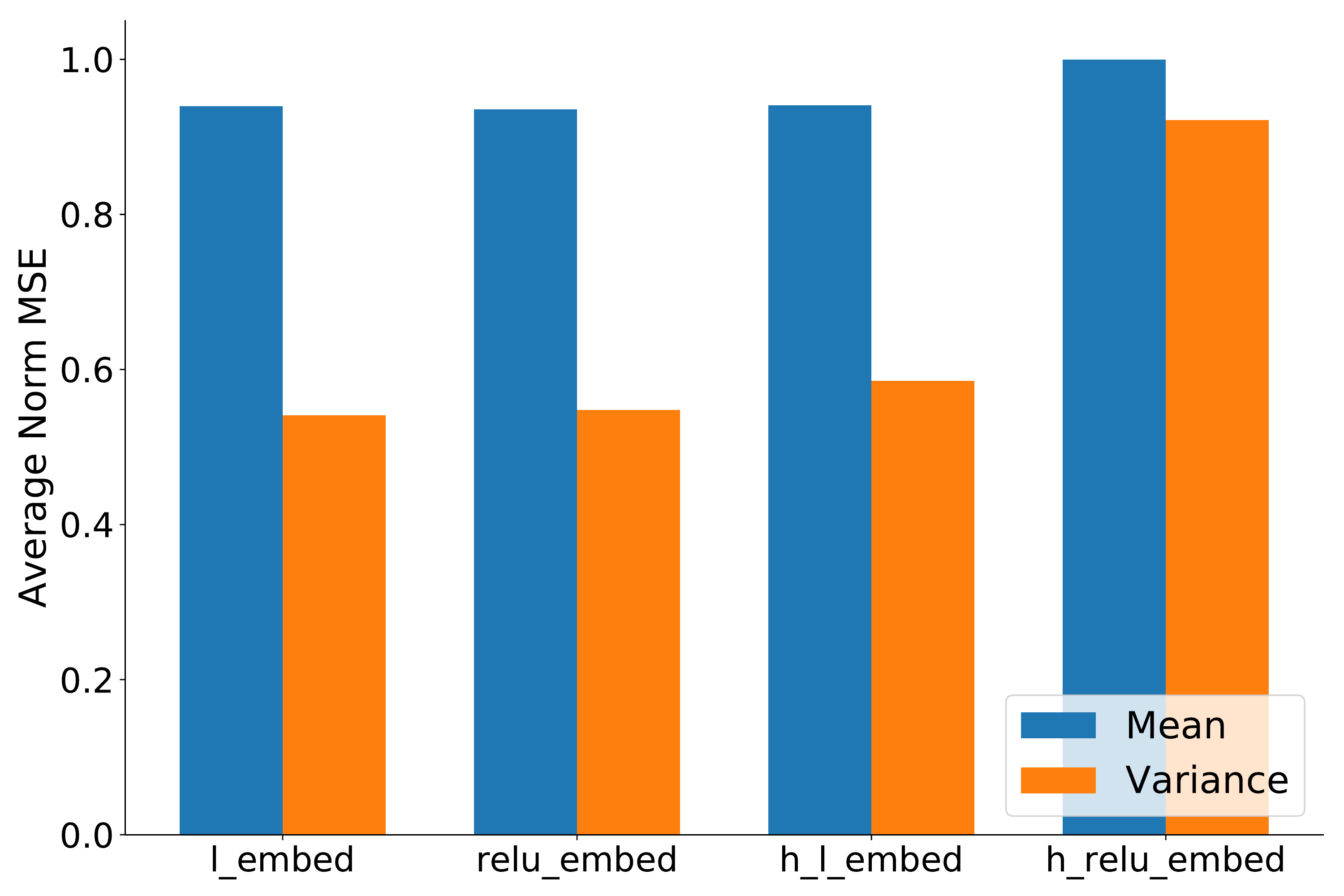}{\textbf{Linear and ReLU Embedding Activations.} If we include Figure~\ref{fig:embedding_2D} it generally appears that linear activations are better than ReLU for the embedding layers.}
	
	\section{Atari@200M}
	\label{sec:additional_atari}
	
	Here we present the results of training {\thenames} on five Atari games for 200 M frames (Figures~\ref{fig:200_games_embedding}---\ref{fig:200_games_scaling}): Asteroids, Atlantis, ChopperCommand, Centipede, and Qbert. We used pretrained networks from the Dopamine package \citep{Dopamine}, trained for 200 M frames. Network configurations are the same as those described in the paper. Each run was trained for 20 M frames and six seeds were run for each experiment. For these results we also included learning curves (rightmost column). These learning curves an average of normalized MSE taken across all evaluation timescales. For each timescale we normalize each of the series by the largest MSE of any series for that timescale. Then for each series we average the normalized MSE across all the timescales. As before, shaded areas indicate max and min.
	
	\newcommand{\imgheighttwo}{1.0in}
	
	\newlength{\tempdima}
	\settoheight{\tempdima}{\includegraphics[width=\linewidth]{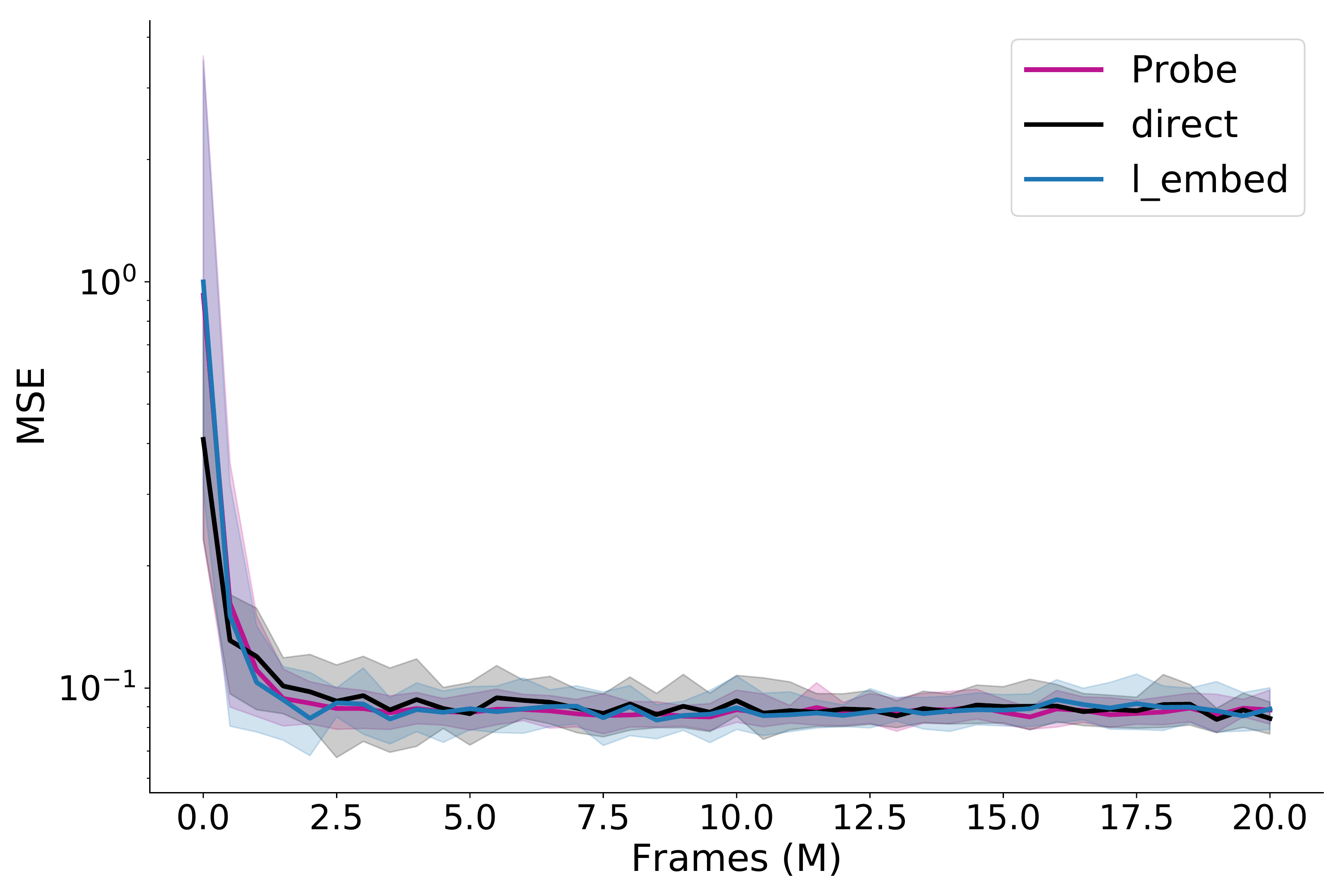}}
	\newcommand{\rowname}[1]{
		\rotatebox{90}{\makebox[\imgheighttwo][c]{\textbf{#1}}}
	}
	
	\newcommand{\twohundredrow}[3][0]{
		\rowname{#2}
		\begin{subfigure}[t]{0.23\linewidth}
			\centering
			\includegraphics[height=\imgheighttwo]{img/p_e/200/#2/#3/by_ts/MSE.pdf}
			\ifthenelse{#1=1}{\caption{MSE}}{}
		\end{subfigure}
		\hfill
		\begin{subfigure}[t]{0.23\linewidth}
			\centering
			\includegraphics[height=\imgheighttwo]{img/p_e/200/#2/#3/by_ts/MSE_bar.pdf}
			\ifthenelse{#1=1}{\caption{Average Norm MSE}}{}
		\end{subfigure}
		\hfill
		\begin{subfigure}[t]{0.23\linewidth}
			\centering
			\includegraphics[height=\imgheighttwo]{img/p_e/200/#2/#3/by_ts/corr.pdf}
			\ifthenelse{#1=1}{\caption{Correlation}}{}
		\end{subfigure}
		\hfill
		\begin{subfigure}[t]{0.23\linewidth}
			\centering
			\includegraphics[height=\imgheighttwo]{img/p_e/200/#2/#3/MSE.pdf}
			\ifthenelse{#1=1}{\caption{Learning Curves}}{}
		\end{subfigure}
	}
	
	\newcommand{\twohundred}[2]{
		\begin{figure*}[t!]
			\centering
			\twohundredrow{Asteroids}{#1}
			\\
			\twohundredrow{Atlantis}{#1}
			\\
			\twohundredrow{Centipede}{#1}
			\\
			\twohundredrow{ChopperCommand}{#1}
			\\
			\twohundredrow[1]{Qbert}{#1}
			\caption{#2}
			\label{fig:200_games_#1}
		\end{figure*}
	}
	
	\twohundred{embedding}{\textbf{Atari@200M: Embedding Comparison.} We find no consistent difference between using the \textit{direct} or \textit{l\_embed} approaches.}
	
	\twohundred{inputs}{\textbf{Atari@200M: Inputs Comparison.} We see that using $\gamma$ as input gives better results at very short time scales than $\tau$, but otherwise $\tau$ is better. Overall, providing both $\gamma$ and $\tau$ provides the best performance. Although, the results in Asteroids are notable exception.}
	\twohundred{dist}{\textbf{Atari@200M: Distribution Comparison.} Populating $\Kset_t$ from the $\tau$ scale is better than from $\gamma$ scale except at very short timescales. Drawing samples from both scales does best overall.}
	\twohundred{scaling}{\textbf{Atari@200M: Scaling Comparison.} As expected scaling the loss generally helped improve performance for shorter timescales, at the cost of performance elsewhere. We note that the error at initialization, which can be see in the learning rate plots, can be much lower without the scaling. This is due to the division by $(1-\gamma)$ used in the scaling networks, which has the effect of amplifying the noise in the initialization. To counter this we tried reducing the initial network weights, $\theta$, by multiplying by $(1-\gamma)$ (\textit{scaled $\delta$ and $\theta$}). This did improve the initial error and matched the loss scaling for performance. This also appeared to reduce variance in some cases.}
	
%
	
\end{document}